\documentclass[10pt,journal,letterpaper,compsoc]{IEEEtran}

\ifCLASSOPTIONcompsoc

\else

\fi

\ifCLASSINFOpdf

\else

\fi

\usepackage{url}

\usepackage{times}
\usepackage{epsfig}
\usepackage{graphicx}
\usepackage{amsmath}
\usepackage{amssymb}
\usepackage{subfigure}
\usepackage{multirow}
\usepackage{color}
\usepackage{multirow,bm}
\usepackage[font=small,labelfont=bf]{caption}
\usepackage{bibspacing}

\DeclareMathOperator*{\argmax}{arg\,max}

\begin{document}

\title{Efficient Activity Detection in Untrimmed Video with Max-Subgraph Search}

\author{Chao~Yeh~Chen and~Kristen~Grauman
\IEEEcompsocitemizethanks{\IEEEcompsocthanksitem C.-Y. Chen is with the Department
of Computer Science, University of Texas at Austin, Texas,
TX, 78712.\protect\\
E-mail: chaoyeh@cs.utexas.edu

E-mail: chaoyeh@cs.utexas.edu
\IEEEcompsocthanksitem K. Grauman is with the Department
of Computer Science, University of Texas at Austin, Texas,
TX, 78712.\protect\\
E-mail: grauman@cs.utexas.edu}
\thanks{}}


\IEEEcompsoctitleabstractindextext{%
\begin{abstract}
We propose an efficient approach for activity detection in video that unifies activity categorization with space-time localization.  The main idea is to pose activity detection as a maximum-weight connected subgraph problem.  Offline, we learn a binary classifier for an activity category using positive video exemplars that are ``trimmed" in time to the activity of interest.  Then, given a novel \emph{untrimmed} video sequence, we decompose it into a 3D array of space-time nodes, which are weighted based on the extent to which their component features support the learned activity model.  To perform detection, we then directly localize instances of the activity by solving for the maximum-weight connected subgraph in the test video's space-time graph.  We show that this detection strategy permits an efficient branch-and-cut solution for the best-scoring---and possibly non-cubically shaped---portion of the video for a given activity classifier.  The upshot is a fast method that can search a broader space of space-time region candidates than was previously practical, which we find often leads to more accurate detection.  We demonstrate the proposed algorithm on four datasets, and we show its speed and accuracy advantages over multiple existing search strategies.
\end{abstract}

}

\maketitle

\IEEEdisplaynotcompsoctitleabstractindextext

\IEEEpeerreviewmaketitle

\section{Introduction}\label{sec:intro}

The activity detection problem entails both \emph{recognizing} and \emph{localizing} categories of activity in an ongoing (meaning ``untrimmed") video sequence.  In other words, a system must not only be able to recognize a learned activity in a new clip; it must also be able to isolate the (potentially small) portion of a long input sequence that contains the activity.  Reliable activity detection would have major practical value for applications such as video indexing, surveillance and security, and video-based human computer interaction.

While the recognition portion of the problem has received increasing attention in recent years, state-of-the-art methods largely assume that the space-time region of interest to be classified has already been identified.  However, for most realistic settings, a system must not only name what it sees, but also partition out the temporal or spatio-temporal extent within which the activity occurs.  The distinction is non-trivial; in order to properly recognize an action, the spatio-temporal extent usually must be known \emph{simultaneously}.

\begin{figure}[t]
\centering
\includegraphics[width=0.45\textwidth]{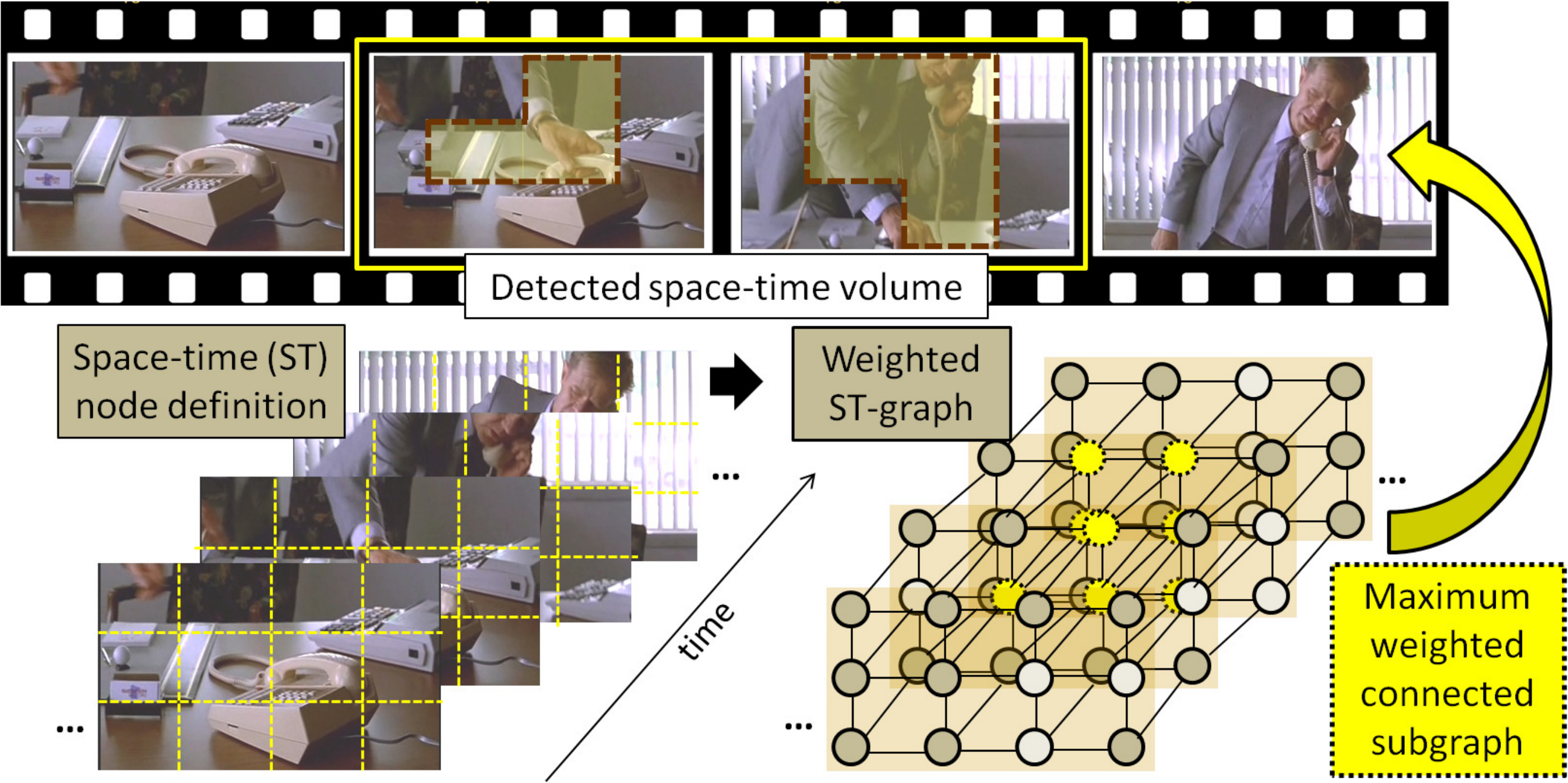}
   \caption{Our approach constructs a space-time video graph, and efficiently finds the subgraph that maximizes an activity classifier's score.  The detection result can take on non-cubic shapes (see dotted shapes in top frames), as demanded by the action.}\label{fig:cartoon}
\end{figure}

To meet this challenge, existing methods tend to separate activity detection into two distinct stages: the first generates space-time candidate regions of interest from the test video, and the second scores each candidate according to how well it matches a given activity model (often a classifier).  Most commonly, candidates are generated either using person-centered tracks~\cite{Moore:1999:trackingrecognition,ramanan-nips2003,Yao:2010:houghvoting,Klaser:2010:humanfocusedactionlocalization} or using exhaustive sliding window search through all frames in the video~\cite{Ke:2005:volumetricfeatures,Duchenne:2009:automaticannotation,Satkin:2010:modelingtemporalextent}.  Both face potential pitfalls.  On the one hand, a method reliant on tracks is sensitive to tracking failures, and by focusing on individual humans in the video, it overlooks surrounding objects that may be discriminative for an activity (e.g., the car a person is approaching).  On the other hand, sliding window search is clearly a substantial computational burden, and its frame-level candidates may be too coarse, causing clutter features to mislead the subsequent classifier.  In both cases, the scope of space-time regions even considered by the classifier is artificially restricted, e.g., to person bounding boxes or a cubic subvolume.

Our goal is to unify the classification and localization components into a single detection procedure.  We propose an efficient approach that exploits top-down activity knowledge to quickly identify the portion of video that maximizes a classifier's score.  In short, it works as follows.  Given a novel video, we construct a 3D graph in which nodes describe local video subregions, and their connectivity is determined by proximity in space and time.  Each node is associated with a learned weight indicating the degree to which its appearance and motion support the action class of interest.  Using this graph structure, we show the detection problem is equivalent to solving a \emph{maximum-weight connected subgraph} problem, meaning to identify the subset of connected nodes whose total weight is maximal.  For our setting, this in turn is reducible to a prize-collecting Steiner tree problem, for which practical branch-and-cut optimization strategies are available.  This means we can efficiently identify both the spatial and temporal region(s) within the sequence that best fit a learned activity model.  See Figure~\ref{fig:cartoon}.

The proposed approach has several important properties.  First, for the specific case where our space-time nodes are individual video frames, the detection solution is equivalent to that of exhaustive sliding window search, yet costs orders of magnitude less search time due to the branch-and-cut solver.  Second, we show how to create more general forms of the graph that permit ``non-cubic" detection regions, and even allow hops across irrelevant frames in time that otherwise might mislead the classifier (e.g., due to a temporary occluding object).  This effectively widens the scope of candidate video regions considered beyond that allowed by any prior methods; the upshot is improved accuracy.  Third, we explore a two-stage search extension that increases the speed of the proposed subgraph search for long videos, and show its generality for detecting multiple activity instances in a single input sequence.  Finally, the method accommodates a fairly wide family of features and classifiers, making it flexible as a general activity detection tool.  To illustrate this flexibility, we devise a novel high-level descriptor amenable to subgraph search that reflects human poses and objects as well as their relative temporal ordering.

We validate the algorithm on four challenging datasets.  The results demonstrate its clear speed and accuracy advantages over both standard sliding window search as well as a state-of-the-art branch-and-bound solution~\cite{Yuan:2009:subvolumesearch}.

\section{Related Work}\label{sec:Related Work}

\begin{figure*}[t]
\centering
\hspace*{-0.4in}
\includegraphics[width=0.7\textwidth]{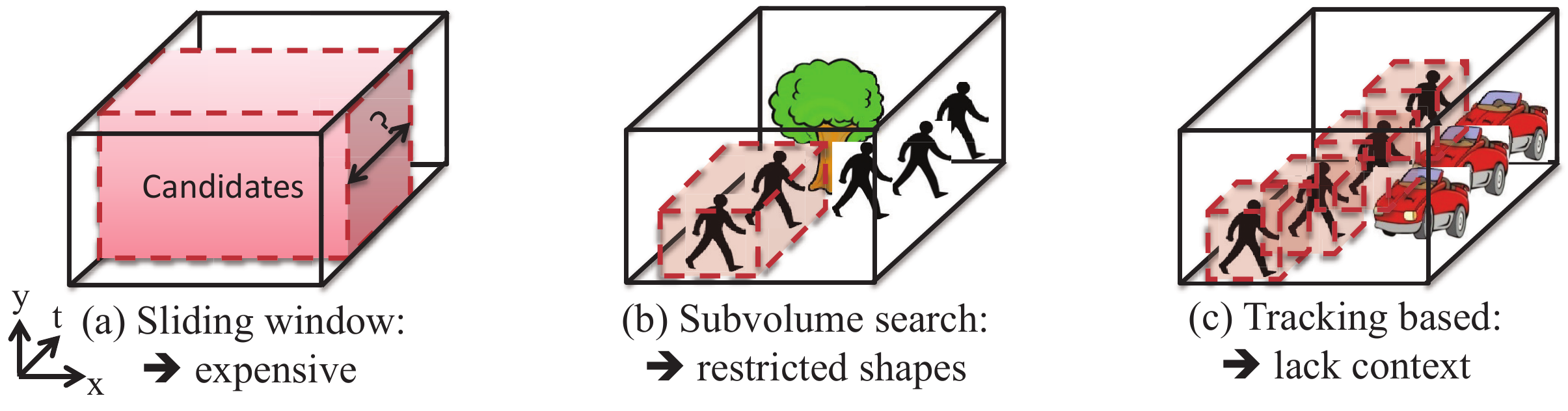}
   \caption{Overview and limitations of standard approaches to activity detection.  (a) A sliding window search is most commonly employed, but it is computationally expensive to search all possible windows.  Furthermore, searching only in the temporal dimension can mislead a classifier when there is substantial clutter or irrelevant features in the scene, i.e., the objects relevant to the activity occupy a small portion of the frame. (b) Efficient subvolume search techniques can greatly improve efficiency and generalize sliding windows to restricted spatial areas, yet existing methods are limited to cubic subvolumes.  This creates problems with the subject of the activity moves within the frame.  (c) A general alternative to sliding window search is to detect activities based on tracked people (or other objects).  This lets the system follow the region of interest as it moves over time and works well for ``person-centric" activities like waving or jumping.  However, the tracking-based approach risks losing important contextual information about an activity (e.g., as depicted here, the activity ``get into car" demands a representation of surrounding objects, not just the tracked person).}
\label{fig:status_quo}
\end{figure*}

We focus our literature review on methods handling action \emph{detection} in video.  There is also a large body of work in activity \emph{recognition} (from either a sequence or a static frame) where one must categorize a clip/frame that is already trimmed to the action of interest.  Representations developed in that work are complementary and could enhance results attainable with our detection scheme.

One class of methods tackles detection by explicitly tracking people, their body parts, and nearby objects (e.g.,~\cite{Moore:1999:trackingrecognition,ramanan-nips2003,Klaser:2010:humanfocusedactionlocalization}).  Tracking ``movers" is particularly relevant for surveillance data where one can assume a static camera.  However, as shown in Figure~\ref{fig:status_quo}(c), relying on tracks can be limiting; it makes the detector sensitive to tracking errors, which are expected in video with large variations in backgrounds or rapidly changing viewpoints (e.g., movies or YouTube video).  Furthermore, while good for activities that are truly person-based---like handwaving or jumping jacks---a representation restricted to person-tracks will suffer when defining elements of the action are external to people in the scene (e.g., the computer screen a person is looking at, or the chair he may sit in).

Conscious of the difficulty of relying on tracks, another class of methods has emerged that instead treats activity classes as learned space-time appearance and motion patterns.  The bag of space-time interest point features model is a good example~\cite{Laptev:2008:hollywood,Laptev:2004:recognizinghumanactions}.  In this case, at detection time the classifier is applied to features falling within candidate subvolumes within the sequence.  Typically the search is done with a sliding window over the entire sequence~\cite{Ke:2005:volumetricfeatures,Duchenne:2009:automaticannotation,Satkin:2010:modelingtemporalextent}, or in combination with person tracks~\cite{Klaser:2010:humanfocusedactionlocalization}.

Given the enormous expense of such an exhaustive search for sliding window method, some recent work explores branch-and-bound solutions to efficiently identify the subvolume that maximizes an additive classifier's output~\cite{Yuan:2009:subvolumesearch,Yu:2011:randomforestfastsearch,Cao:2010:crossdataset}.  This approach offers fast detection and can localize activities in both space and time, whereas sliding windows localize only in the temporal dimension.  However, as shown in Figure~\ref{fig:status_quo}(b), in contrast to our approach, existing branch-and-bound methods are restricted to searching over \emph{cubic} subvolumes in the video; that limits detections to cases where the subject of the activity does not change its spatial position much over time.  Our results demonstrate the value of the more general detection shapes allowed by our method.

An alternative way to avoid exhaustive search is through voting algorithms.  Recent work explores ways to combine person-centric tracks or pre-classified sequences with a Hough voting stage to refine the localization~\cite{Yao:2010:houghvoting,mikol,sunil-iccv2013}, or to use voting to generate candidate frames for merging~\cite{willems}.  Like any voting method, such approaches risk being sensitive to noisy background descriptors that also cast votes, and in particular will have ambiguity for actions with periodicity.  Furthermore, in contrast to our algorithm, they cannot guarantee to return the maximum scoring space-time region for a classifier.

Rather than pose a detection task, the multi-class recognition approach of~\cite{Hoai:2011:jointsegmentation} uses dynamic programming to select the temporal boundaries per action.  Like our technique it jointly considers recognition and segmentation.  However, unlike our method, it localizes only in the temporal dimension, assumes a multi-class objective where all parts of the sequence will belong to some pre-trained category (thus requiring one to learn a ``background" activity class), and cannot detect multiple activities occurring at the same time.

The branch-and-cut algorithm we use to optimize the subgraph has also been explored for object segmentation in static images~\cite{Sudheendar:2011:maxsubgraphdetection}.  In contrast, our approach addresses activity detection, and we explore novel graph structures relevant for video data.

This article extends our earlier conference paper~\cite{chaoyeh-cvpr2012}.  The main new additions are (1) an extension to the method to permit efficient spatially localized search even over long sequences, (2) an extension to the method to detect multiple instances of an activity in a single sequence, (3) new results on a fourth dataset specifically designed for the detection task (THUMOS 2014~\cite{THUMOS14}), (4) new qualitative results, and (5) new figures to better present our ideas and approach.

\section{Approach}\label{sec:Algorithm Sketch}

\begin{figure*}[t]
\centering
\includegraphics[width=\textwidth]{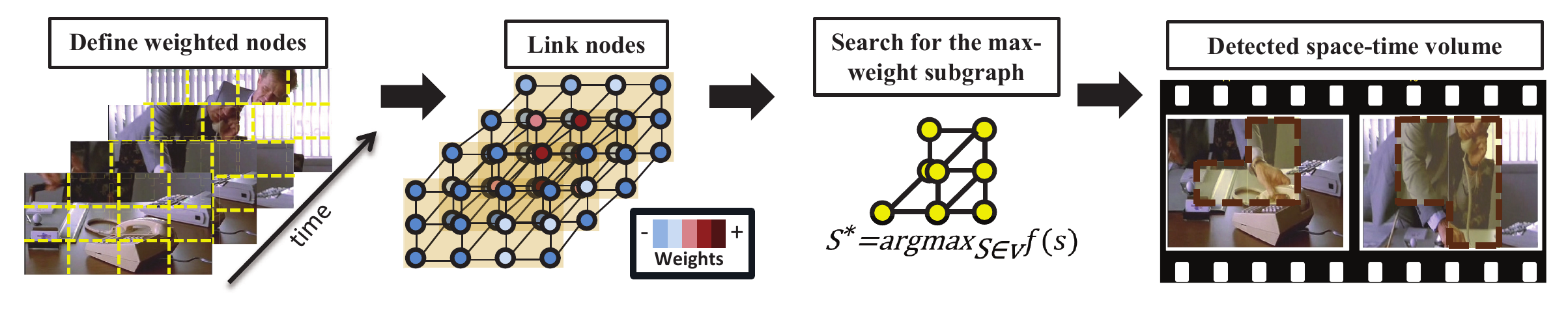}
   \caption{Pipeline of our max-subgraph search approach to activity detection.  To detect an action in a video, we first divide the test video into a 3D array of spatio-temporal volumes.  These are the graph nodes, weighted by their respective features' classifier scores (Sec.~\ref{sec:objective}).  The edges in the graph connect nodes that are adjacent in space-time (Sec.~\ref{sec:graph}), with optional additional connectivity between more distant nodes in order to tolerate occlusions or noisy features (i.e., as defined for the T-Jump variant).
 Then we search for the maximum weighted subgraph using an efficient branch-and-cut solution (Sec.~\ref{sec:search}).  Finally, the resulting subgraph yields the subset of connected space-time nodes where the action most likely appears.  This is our detection result, as shown backprojected into the original video in the rightmost image above.}
\label{fig:framework}
\end{figure*}

Our approach first trains a detector using a binary classifier and training examples where the action's temporal extent is known.  Then, given test sequences for which we have no knowledge of the start and end of the activity, it returns the subsequence (and optionally, the spatial regions of interest) that maximizes the classifier score.  This works by creating a space-time graph over the entire test sequence, where each node is a space-time cube, and the cubes are linked according to their proximity in space and time.  Each node is weighted by a positive or negative value indicating its features' contribution to the classifier's score.  Thus, the subsequence for which the detector would yield the maximal score is equivalent to the maximum weight connected subgraph.  This subgraph can be efficiently computed using an existing branch-and-cut algorithm, thereby finding the optimal solution without exhaustive search through all possible sets of connected nodes.

We first define the classifiers accommodated by our method (Sec.~\ref{sec:objective}), and the features we use (Sec.~\ref{sec:features}).  Then we describe how the graphs are constructed (Sec.~\ref{sec:graph}); we introduce variants of the node structure and linking strategy that allow us to capture different granularities at detection time.  Next, we briefly explain the maximum subgraph problem and branch-and-cut search (Sec.~\ref{sec:search}). Finally, we devise two extensions of our basic framework that can deal with spatio-temporal detection even in long videos (Sec.~\ref{sec:two_stage}) and detection of multiple instances in a single sequence (Sec.~\ref{sec:multiple_instance_detection}).

\subsection{Detector Training and Objective}\label{sec:objective}

We are given labeled training instances of the activity of interest, and train a binary classifier $f: S \rightarrow \mathbb{R}$ to distinguish positive instances from all other action categories.  This classifier can score any subvolume $S$ of a novel video according to how well it agrees with the learned activity.  To perform activity detection, the goal is to determine the subvolume in a new sequence $Q$ that maximizes the score
\begin{equation}
S^\ast = \argmax_{S \in Q}~~f(S).\label{eqn:maximize}
\end{equation}
If we were to restrict the subvolume in the spatial dimensions to encompass the entire frame, then $S^\ast$ would correspond to the output of an exhaustive sliding window detector.  More generally, the optimal subvolume $S^\ast$ is the set of contiguous voxels of arbitrary shape in $Q$ that returns the highest classifier score.

Our approach requires the classifier to satisfy two properties.  First, it must be able to score an arbitrarily shaped set of voxels.  Second, it must be defined such that features computed within local space-time regions of the video can be combined \emph{additively} to obtain the classifier response for a larger region.  The latter is necessary so that we can decompose the classifier response across the nodes of the space-time graph, and thereby associate a single weight with each node.  Suitable additive classifiers include linear support vector machines (SVM), boosted classifiers, or Naive Bayes classifiers computed with localized space-time features, as well as certain non-linear SVMs~\cite{vedaldi-kernels}.

Our results use a linear SVM with histograms (bags) of quantized space-time descriptors.  The bag-of-features (BoF) representation has been explored in a number of recent activity recognition methods (e.g.,~\cite{Laptev:2008:hollywood,kalser:2008:HoG3D,niebles}), and, despite its simplicity, offers very competitive results.  We consider BoF's computed over two forms of local descriptors.  The first consists of low-level
histograms of oriented gradients and flow computed at space-time interest points; the second consists of a novel high-level descriptor that encodes the relative layout of detected humans, objects, and poses.  Both descriptors are detailed below in Sec.~\ref{sec:features}.

In either case, we compute a vocabulary of $K$ visual words by quantizing a corpus of features from the training images.  A video subvolume with $N$ local features is initially described by the set $S = \{(\bm{x}_i, \bm{v}_i)\}_{i=1}^N$, where each $\bm{x}_i = (x_i,y_i,t_i)$ refers to the 3D feature position in space and time, and $\bm{v}_i$ is the associated local descriptor.  Then the subvolume is converted to a $K$-dimensional BoF histogram $h(S)$ by mapping each $\bm{v}_i$ to its respective visual word $c_i$, and tallying the word counts over all $N$ features.

We use the training instances to learn a linear SVM, which means the resulting scoring function has the form:
\begin{equation}
f(S) = \beta + \sum_i \alpha_i \langle h(S), h(S_i) \rangle,
\end{equation}
where $i$ indexes the training examples, and $\alpha$, $\beta$ denote the learned weights and bias.  This can be rewritten as a sum over the contributions of each feature.  Let $h^j(S)$ denote the $j$-th bin count for histogram $h(S)$.  The $j$-th word is associated with a weight
\begin{equation}
w^j = \sum_i \alpha_i h^j(S_i),
\end{equation}
for $j=1,\dots,K$.  Thus the classifier response for a subvolume $S$ is:
\begin{align}
  f(S) &= \beta + \sum_{j=1}^K w^j h^j(S) \\
  &= \beta + \sum_{i=1}^{N} w^{c_i},
\label{eqn:fi}
\end{align}
where again $c_i$ is the index of the visual word that feature $\bm{v}_i$ maps to, $c_i \in [1,K]$.  By writing the score of a subvolume as the sum of its $N$ features' ``word weights", we now have a way to associate each local descriptor occurrence with a single weight---its contribution to the total classifier score.\footnote{The bias term $\beta$ can be ignored for the purpose of maximizing $f(S)$.}  This same property of linear SVMs is used in~\cite{Lampert:2008:efficientsubwindow} to enable efficient subwindow search for object detection, whereas we exploit it to score non-cubic subvolumes in video for action detection.

We stress that our method is not limited to linear SVMs; alternative additive classifiers with the properties described above are also permitted.  Our experiments in Sec.~\ref{sec:Experiment Result} focus on linear SVMs due to their efficacy.  We have also successfully implemented the framework using others, e.g., Naive Bayes, with the same input features.  The results are sound, however across the board we find that classifier is less effective than the SVM for our task.

Furthermore, while the additive requirement does lead to an orderless bag-of-features representation, it is still possible to encode temporal ordering into the approach depending on how the local descriptors are extracted.  For example, in Sec.~\ref{sec:high} we provide one way to record the space-time layout of neighboring objects into high-level visual words.

\subsection{Localized Space-Time Features}\label{sec:features}

We consider two forms of localized descriptors for the $\bm{v}_i$ vectors above: a conventional low-level gradient-based feature, and a novel high-level feature.

\subsubsection{Low-level Descriptors}

For low-level features, we employ an array of widely used local video descriptors from the literature.  In general, they capture the texture and motion within localized space-time volumes, either at interest points or dense positions within the video.  In particular, we use histograms of oriented gradients (HoG) and histograms of optical flow (HoF) computed in local space-time cubes~\cite{Laptev:2008:hollywood,kalser:2008:HoG3D}.  The local cubes are centered at either 3D Harris interest points~\cite{stip} or densely sampled.  These descriptors capture the appearance and motion in the video, and their locality lends robustness to occlusions.   We also incorporate dense trajectory~\cite{wang:2011:dense-trajectory} and motion boundary histogram (MBH)~\cite{oneata:2013:mbh} features in a bag-of-features representation.  We refer the reader to the original papers about the descriptors for more details.

As is typical in visual recognition, we can expect better accuracy as a function of the greater the variety and complementarity of the features we use, but with some tradeoff in computational cost.  Specifically, the main influence the features will have on our method's complexity is their density in the video; while their density will not at all affect the node structure (cf.~Sec.~\ref{sec:graph}), it will dictate how many visual word mappings must be computed.  In Sec.~\ref{sec:Experiment Result} we provide more discussion about how we select among these descriptors for different datasets; in short, our selection is largely based on empirical findings from previous work about which are best suited.

\begin{figure}[t]
\centering
\includegraphics[width=0.5\textwidth]{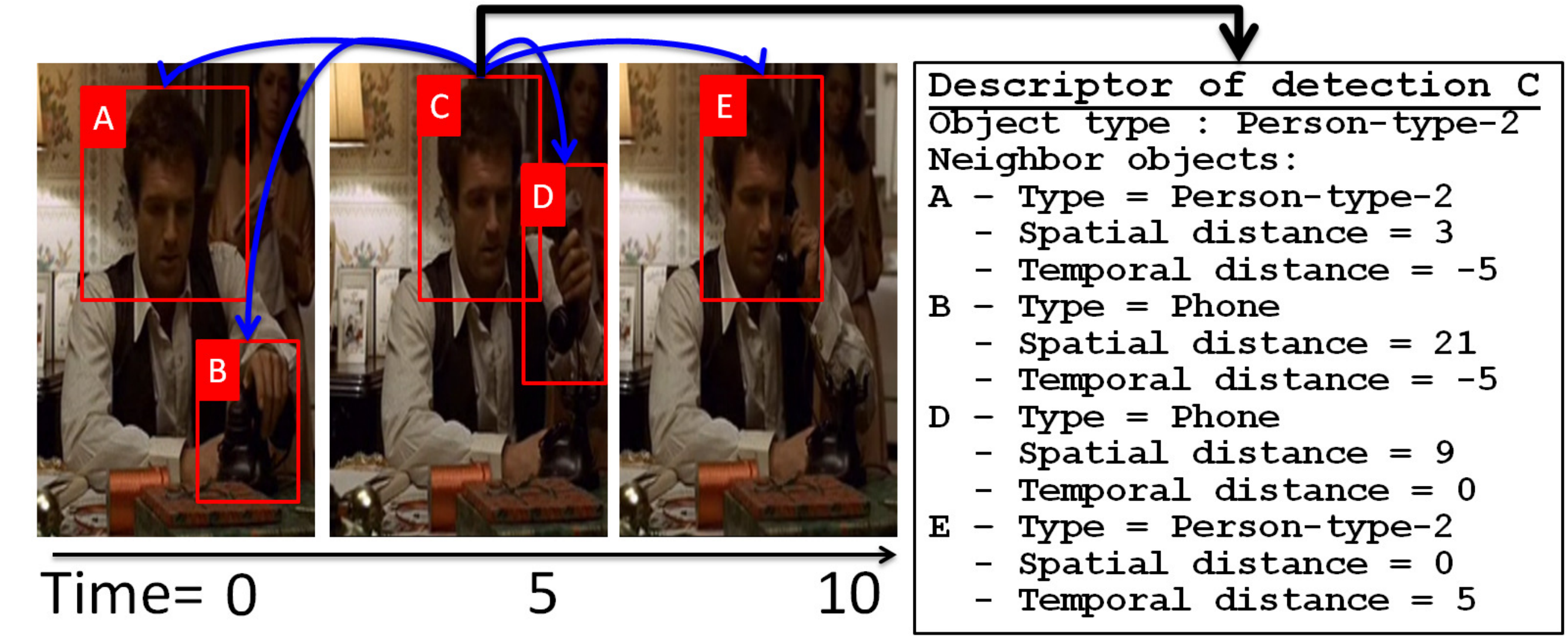}
   \caption{Schematic of the data comprising our high-level descriptors.  After detecting people and other objects in the video frames, we form semi-local neighborhoods around each detected object that summarize the space-time layout of other nearby detections.  To map those neighborhoods into discrete and discriminative visual words, we apply a random forest trained for the action labels (Sec.~\ref{sec:high}).   Here, the left images depict the detected objects surrounding the person detected in bounding box C in the center frame.  The right text box displays the information exposed to the random forest feature quantizer, in terms of the neighboring detections and their relative spatial and temporal distance from that person box C.}
\label{fig:high_feature_concept}
\end{figure}

\subsubsection{High-level Descriptors}\label{sec:high}

We introduce a novel descriptor for an alternative high-level representation.  While low-level gradient features are effective for activities defined by gestures and movement (e.g., running  vs.~diving), many interesting actions are likely better defined in terms of the semantic \emph{interactions} between people and objects~\cite{gupta-pami2009,desai-2010,prest-pami2012}.  For example, ``answering phone'' should be compactly describable in terms of a person, a reach, a grasp of the receiver, etc.

To this end, we compose a descriptor that encodes the objects and poses occurring in a space-time neighborhood.  First, we run a bank of object detectors~\cite{Felzenswalb10} and a bank of mid-level ``poselet'' detectors~\cite{Poselets09} on all frames.  To capture human \emph{pose}, we categorize each detected person into one of $P=15$ ``person types''.  These types are discovered from person detection windows in the training data: for each person window we create a histogram of the poselet activations that overlap it, and then quantize the space of all such histograms with $k$-means to provide $P$ discrete types.  Each reflects a coarse pose---for example, a seated person may cause upper body poselets to fire, whereas a hugging person would trigger poselets from the back.

Given the sparse set of bounding box object detections in a test sequence, we form one neighborhood descriptor per box.  This descriptor reflects (1) the type of detector (e.g., person type \#3, car) that fired at that position, (2) the distribution of object/person types that also fired within a 50-frame temporal window of it, and (3) their relative space-time distances.  See Figure~\ref{fig:high_feature_concept}.

To quantize this complex space into discriminative high-level ``words'', we devise a random forest technique.   When training the random forest, we choose spatial distance thresholds, temporal distance thresholds, and object types to parameterize semantic questions that split the raw descriptor inputs so as to reduce action label entropy.  Each training and testing descriptor is then assigned a visual word according to the indices of the leaf nodes it reaches when traversing each tree in the forest.  Essentially, this reduces each rich neighborhood of space-time object relationships to a single quantized descriptor, i.e., a single index $c_i$ in Eqn.~\ref{eqn:fi}.

In contrast to the low-level features, this descriptor encodes space-time ordering, demonstrating that our max-subgraph scheme is not limited to pure bag-of-words representations.   Furthermore, it leads to faster node weight computations, since the number of detected objects is typically much fewer than the number of space-time interest points.

\subsection{Definition of the Space-Time Graph}\label{sec:graph}

So far we have defined the training procedure and features we use.  Now we describe how we construct a space-time graph $G = (V,E)$ for a novel test video, where $V$ is a set of vertices (nodes) and $E$ is a set of edges.  Recall that a test video is ``untrimmed", meaning that we have no prior knowledge about where an action(s) starts or ends in either the spatial or temporal dimensions.  Our detector will exploit the graph to efficiently identify the most likely occurrences of a given activity.  We present two variants each for the node and link structures, as follows.

\subsubsection{Node Structure}

Each node in the graph is a set of contiguous voxels within the video.  In principle, the smallest possible node would be a pixel, and the largest possible node would be the full test sequence.  What, then, should be the scope of an individual node?  The factors to consider are (1) the granularity of detection that is desired (i.e., whether the detector should predict only when the action starts and ends, or whether it should also estimate the spatial localization), and (2) the allowable computational cost.  Note that nodes larger than individual voxels or frames are favorable not only for computational efficiency, but also to aggregate neighborhood statistics to give better support when the classifier considers that region for inclusion.

\begin{figure}[t]
\centering
\begin{tabular}{cc}
\subfigure[Temporal only (T)]{
\centering
\includegraphics[width=3cm]{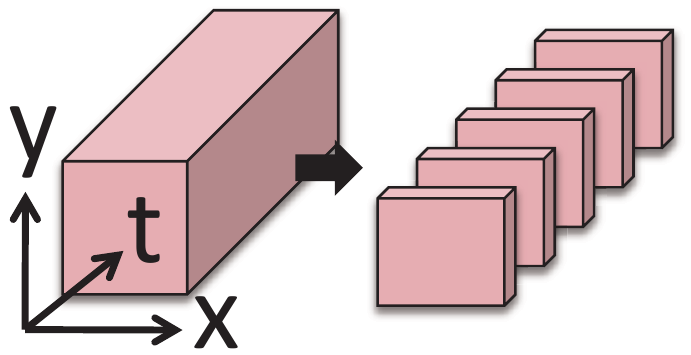}}\hspace*{0.2in}
\subfigure[Spatio-temporal (ST)]{
\centering
\includegraphics[width=3cm]{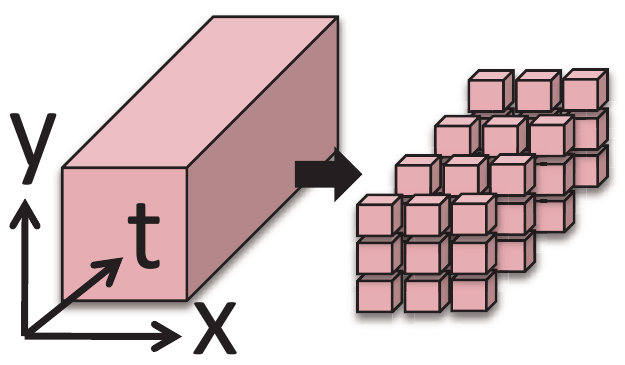}}
\end{tabular}
\caption{The two node structures we consider.  (a) A \emph{temporal only} graph simply breaks the video into slabs of frames.  Max subgraph search on this graph is equivalent to sliding window in terms of results, but is faster.  (b) \emph{Spatio-temporal} graphs further break the frames into spatial cubes, allowing both spatial and temporal localization of the activity in irregular subvolume shapes, at the cost of a denser input graph.}
\label{fig:node_formulation}
\end{figure}

With this in mind, we consider two possible node structures.  The first breaks the video into frame-level slabs, such that each node is a sequence of $F$ consecutive frames.  The second breaks the video into a grid of $H \times W \times F$ space-time cubes.  In all our results, we set $F=5$ or $10$, and let $H$ and $W$ be  $\frac{1}{3}$ of the frame dimensions.\footnote{Rather than space-time cubes, one could consider using space-time \emph{segments} from a bottom-up grouping algorithm.  This would have some potential advantages, including finer-grained localization.  However, our preliminary attempts indicated that the regular grid nodes are preferable to segments in practice, for both accuracy and speed.  That is because (1) the irregularly shaped segment nodes lead to dense adjacency structures, hurting run-time, and (2) the difficulty in producing quality supervoxels makes it easy to over/under-segment.}  See Figure~\ref{fig:node_formulation}.  At detection time, the two forms yield a \emph{temporal subgraph} (\textbf{T-Subgraph}) and \emph{spatio-temporal subgraph} (\textbf{ST-Subgraph}), respectively.  Note that a T-Subgraph will be equivalent to a sliding window search result with a frame step size of $F$.  In contrast, a ST-Subgraph will allow irregular, non-cubic detection results.  See the first and last images in Figure~\ref{fig:shapes}.

After building a graph with either node structure for a test video, we compute the weight for each node $\mathrm{v}$:
\begin{equation}
\omega(\mathrm{v}) = \sum_{\bm{x}_j \in \mathrm{v}} w^{c_j},
\end{equation}
where $\bm{x}_j$ is the 3D coordinate of the $j$-th local descriptor falling within node $\mathrm{v} \in V$, and $c_j$ is its quantized feature index.
We assign the features from Sec.~\ref{sec:features} to their respective graph nodes as follows.  For the case of low-level features,
$\bm{x}_j$ is the space-time interest point position.  For the case of high-level features, $\bm{x}_j$ is the center of the originating object detection window.  In either case, a feature is claimed by the space-time node containing its central position.

Intuitively, nodes with high positive weights indicate that the activity covers that space-time region, while nodes with negative weights indicate the absence of the activity.

\begin{figure}[t]
\centering
\begin{tabular}{cc}
\subfigure[Neighbors only]{
\centering
\includegraphics[width=3.5cm]{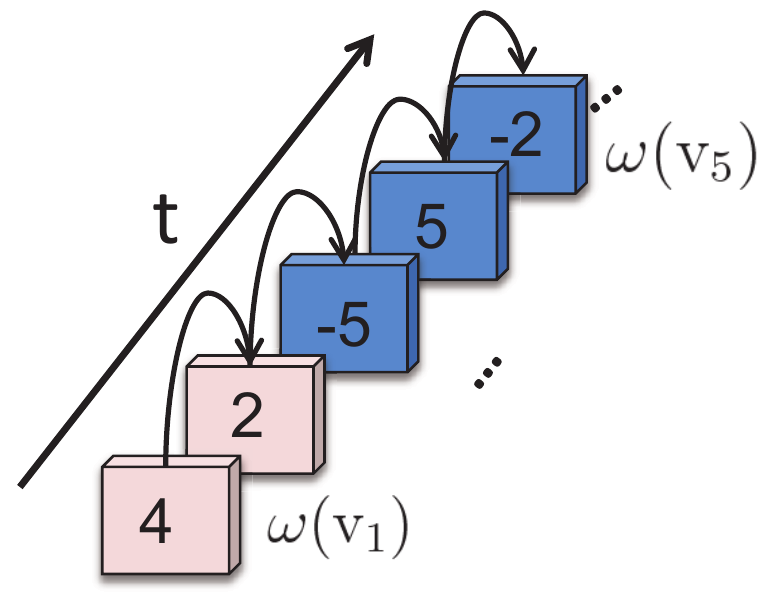}}\hspace*{0.2in}
\subfigure[Neighbors + ``Jump"]{
\centering
\includegraphics[width=3cm]{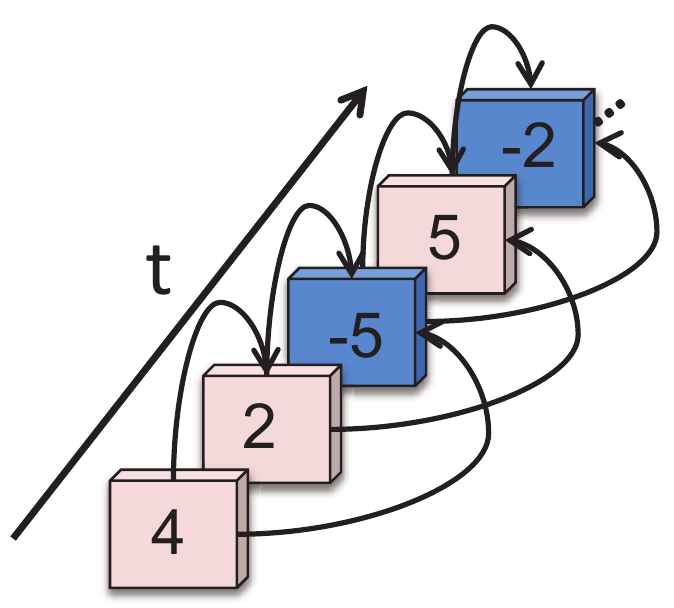}}
\end{tabular}
\caption{The two linking strategies we consider.  (a) The \emph{neighbors only} graph links temporally adjacent (shown here) and optionally spatially adjacent (not shown) nodes.  (b) The \emph{temporal ``jump"} linking strategy also incorporates edges between non-adjacent nodes, so that the output detection can realize a good connected detection result in spite of intermittent noisy/occlusion features on certain nodes.  Here, the numbers shown on nodes indicate weights; white nodes indicate those that would be selected under either linking strategy (see text).}
\label{fig:connect_structure}
\end{figure}

\subsubsection{Linking Strategies}

The connectivity between nodes also affects both the shape of candidate subvolumes and the cost of subgraph search.  We explore two strategies.  In the first, we link only those neighboring nodes that are temporally (and spatially, for the ST node structure) adjacent (see Figure~\ref{fig:connect_structure} (a)).  In the second, we additionally link nodes that are within the first two temporal neighbors (see Figure~\ref{fig:connect_structure} (b)); we call this variant \textbf{T-Jump-Subgraph}. Since at test time we will seek a maximum scoring \emph{connected} subgraph, the former requires detection subvolumes to be strictly contiguous in time (and thus equates to the options available to a sliding window), while the latter allows subvolumes that ``jump" over an adjacent neighbor in time.

By allowing jumps, we can ignore misleading features that may interrupt an otherwise good instance of an action.  For example, Figure~\ref{fig:connect_structure} depicts some temporal nodes and their associated weights $\omega(\mathrm{v}_i)$'s, under either connectivity scheme.  The max subgraph \emph{without} jumps in (a) is the first two nodes only; in contrast, for the same node weights, the max subgraph \emph{with} jumps in (b) extends to include the fourth node, yielding a higher weight subgraph (4+2+5 vs.~4+2).  This can be useful when the skipped node(s) contain noisy features, such as an object temporarily blocking the person performing the activity.  Like the space-time nodes presented above, the use of temporal jumps further expands the space of candidate subvolumes our method can search, at some additional computational cost.

\subsection{Searching for the Maximum Weight Subgraph}\label{sec:search}

Having defined the graph constructed on an untrimmed test sequence, we are ready to describe the detection procedure to maximize $f(S)$ in Eqn.~\ref{eqn:maximize}. Our detection objective is an instance of the maximum-weight connected subgraph problem (MWCS):  \emph{Given a connected undirected, vertex-weighted graph $G = (V, E)$ with weights $\omega: V \rightarrow \mathbb{R}$, find a connected subgraph $T = (V_T \subseteq V, E_T \subseteq E)$ of $G$, that maximizes the score $W(T) = \sum_{\mathrm{v} \in V_T} \omega(\mathrm{v})$.} The best-scoring subgraph is the subvolume in the video most likely to encompass the activity of interest.  That is the output of our approach.  In Sec.~\ref{sec:multiple_instance_detection} we explain how we iteratively apply the subgraph search procedure to retrieve multiple detections in the same video.

With both positive and negative weights, the problem is NP-complete~\cite{ideker}; an exhaustive search would enumerate and score all possible subsets of connected nodes.  However, MWCS can be transformed into an instance of the prize-collecting Steiner tree problem (PCST)~\cite{Dittrich:2008:maxsubgraph} which has the same graph structure as original MWCS and vertex profits $p>0$ and edge costs $c>0$. This MWCS is solvable by transforming the graph into a directed graph and formulating an integer linear programming (ILP) problem with binary variables for every vertex and edge. Then by relaxing the integrality requirement, the problem can be solved with linear programming using a branch-and-cut algorithm (see~\cite{ljubic}). This method gives optimal solutions and is very efficient in practice for the space-time graphs in our setting.

\subsection{Two Stage Spatio-temporal Detection}\label{sec:two_stage}

Next we describe an extension to the framework that further improves efficiency of spatio-temporal detections, at some loss in search completeness. Basically this extension offers a way to further scale-up our detection strategy for long input videos.  It is relevant in the spatio-temporal detection variant of our method (cf.~Fig.~\ref{fig:node_formulation}(b)), not the temporal-only variant (cf.~Fig.~\ref{fig:node_formulation}(a)).  The fine-grained space-time detection offered by the ST-Subgraph comes from its greater number of nodes and denser connectivity.  In particular, in terms of the number of edges as a function of the number of frames, for temporal-only graph, one more temporal node will add one more edge, as for spatio-temporal graph, one more temporal node will add number of edges quadratically to the spatial nodes. Thus, to detect the activity efficiently without reducing the granularity of search scope, we consider how a modest sacrifice on detection accuracy (i.e., giving up the exhaustive search equivalency promised so far) can yield a significantly larger detection speed-up.

\begin{figure}[t]
\centering
\hspace*{-0.2in}
\includegraphics[width=0.5\textwidth]{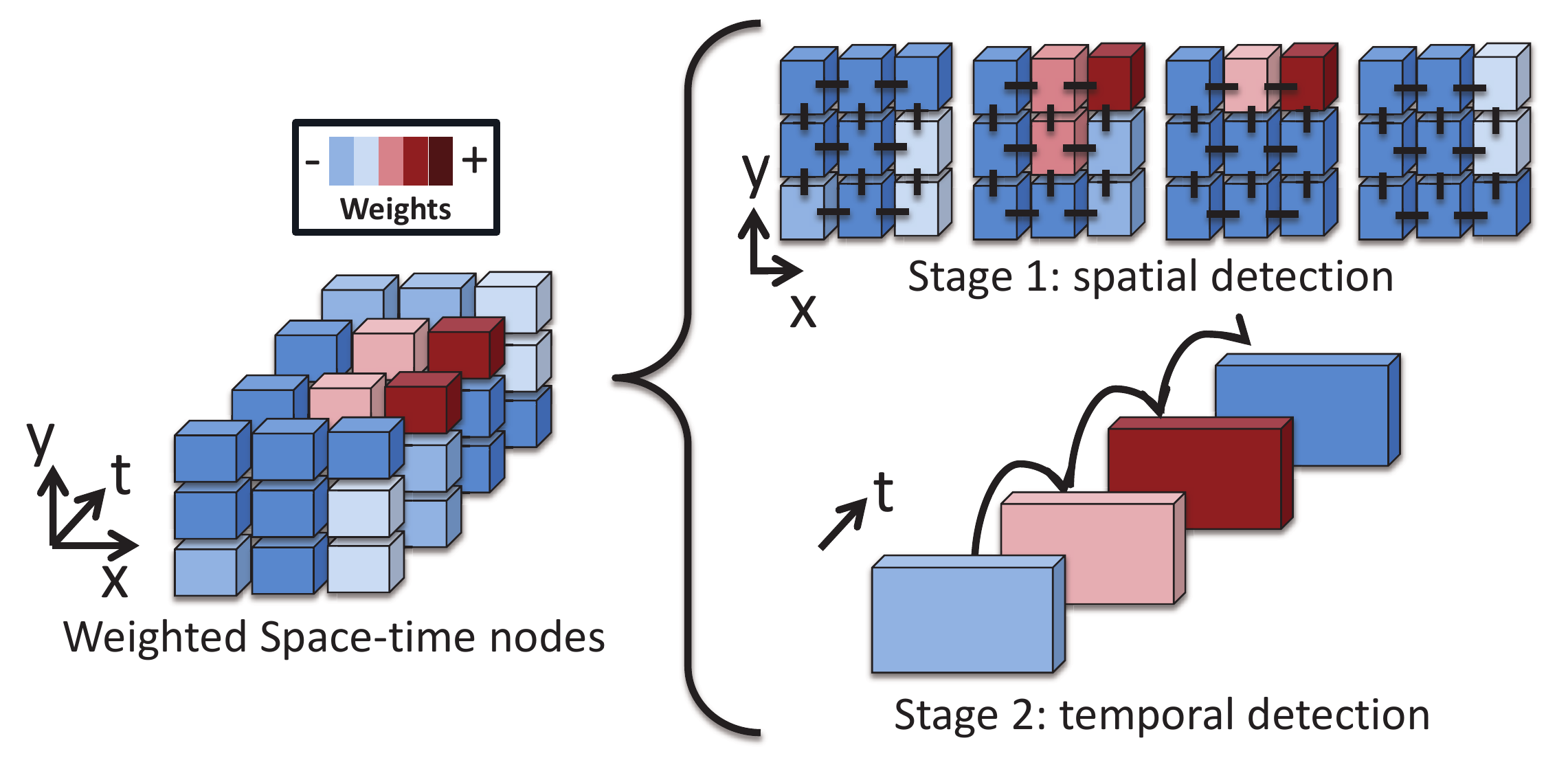}
   \caption{Our two stage subgraph search approximates the ST-Subgraph search, allowing efficient spatio-temporal detection even with long test sequences.  First we extract the standard space-time cuboid nodes (left).  Then, we generate a series of simpler graphs in time (stage 1, top right), and solve for the maximum connected subgraph in each one.  This yields a detection region and score for each simpler graph.  Finally, we create a graph based on temporal nodes only, which are weighted by the output scores of the previous stage (stage 2, bottom right).  The nodes selected in both stages serve as the final output.  Best viewed in color.}
\label{fig:two_stage}
\end{figure}

To this end, we propose a hierarchical \emph{bottom-up} two stage strategy for the space-time search setting. The basic idea is to first perform space-time detections in each temporal slab, and then propagate those detection results up to a second level of processing that performs temporal detection across the slabs.  See Figure~\ref{fig:two_stage}.

Given a test video, we divide the video into spatio-temporal nodes (as depicted in Fig.~\ref{fig:two_stage}, left) and compute their weights as described in Sec.~\ref{sec:graph}.  Next, we search for the best detection volume in two stage: (1) a spatial detection stage and (2) a temporal detection stage.  For the spatial detection stage, we connect nodes in the same temporal slice into a 2D connected weighted graph (see Fig.~\ref{fig:two_stage}, top right).  This yields a series of graphs, each of which has nodes representing the features in different spatial positions in the respective temporal slab.  We then apply the subgraph search procedure from Sec.~\ref{sec:search} to find the maximum weighted connected subgraph in each slab.  Next, the detection score for each 2D subgraph is used to represent the weight of each temporal slab, and these slabs are connected into a 1D temporal graph (see Fig.~\ref{fig:two_stage}, bottom right).  Finally, we find the maximum weighted subgraph along the temporal dimension to obtain the detection output.  The spatio-temporal detection result is determined by set of spatial-temporal nodes in the 2D max-subgraph that are also selected in 1D max-subgraph.

This hierarchical process reduces the computational cost by dividing the original 3D graph structure into a 2D$+$1D graph structure. Note, however, that the detection result from two-stage subgraph search may differ from that returned by the original ST-Subgraph.  Whereas the ST-Subgraph is guaranteed to return the same result an exhaustive search over connected subgraphs, in this modified two-stage procedure, the temporal connection between nodes is always reduced to one edge (vs. nine edges for the original ST-Subgraph).  However, the two-stage search process still provides broader searching scope than the simpler T-Subgraph structure.

In practice, when the length of testing video clip is over 1,000 frames, the two-stage subgraph would be preferred over ST-subgraph for efficient spatial temporal localization. Also, the two-stage subgraph is an approximation of ST-subgraph, if the feature is too noisy, the two-stage subgraph may provide lower accuracy since it ignores many edges when computing the maximum weighted subgraph.

\subsection{Detecting Multiple Activity Instances}\label{sec:multiple_instance_detection}

Thus far, we have described detection in terms of localizing the single space-time region most likely to contain the activity of interest, In particular, the max-subgraph search returns the subvolume for which the trained classifier would score most highly out of all possible subvolumes.  To address the scenario where the novel test sequence may contain \emph{multiple} instances of the activity, and/or to provide multiple confidence-rated hypotheses for the detection output, we extend the max-subgraph search technique as follows.

To detect multiple instances, the main idea is to iteratively run the max-subgraph procedure on adjusted versions of the original input graph, each time adjusting the graph to reflect the most recent detection.  The most straightforward approach to modifying the graph would be to take all the nodes selected for the most recent detection and re-weight each one to $-\infty$.  Doing so is equivalent to removing those nodes, and it would force the next search iteration to choose other nodes for its next hypothesis.  This approach has shortcomings in practice, however.  While the max-subgraph output from the first detection is optimal in terms of the classifier and features chosen, it need not be perfect in terms of localizing the actual activity.  So, flattening nodes to have weight $-\infty$ leads to fragmented secondary detections.

Therefore, we instead downweight those nodes already involved in a detection, but we do not remove them from the graph entirely.  Specifically, each node is re-weighted to 0, as determined empirically on validation data.  In this way, the modified graph coming into the next iteration of the max-subgraph computation will favor finding new high-scoring detections, but may still partially re-use portions of the previous detection(s).

The effect of this process is roughly analogous to standard non-maximum suppression (NMS) as applied in object/action detection with sliding windows.  With sliding windows, any window with a positive classifier score could be reported as a detection output.  However, many windows with positive scores overlap highly with others, and are actually covering the same object/action instance.  To reduce redundant detections, NMS is used to select a single representative output window among a group that highly overlaps.  A key parameter that determines the behavior of NMS is the threshold for overlap between detections: candidate windows overlapping with the selected window by more than the selected threshold are not added to the detection output.  When the threshold is high, one generates more detection outputs at the risk of redundancy. The re-weighting value applied to nodes in our graphs is analogous to that threshold.  A NMS threshold of 0 in traditional sliding windows would correspond to a re-weighting value of $-\infty$ in our setup; a higher NMS threshold corresponds to a higher re-weighting value, allowing some overlap in output detections.

\begin{figure}[t]
\includegraphics[width=0.5\textwidth]{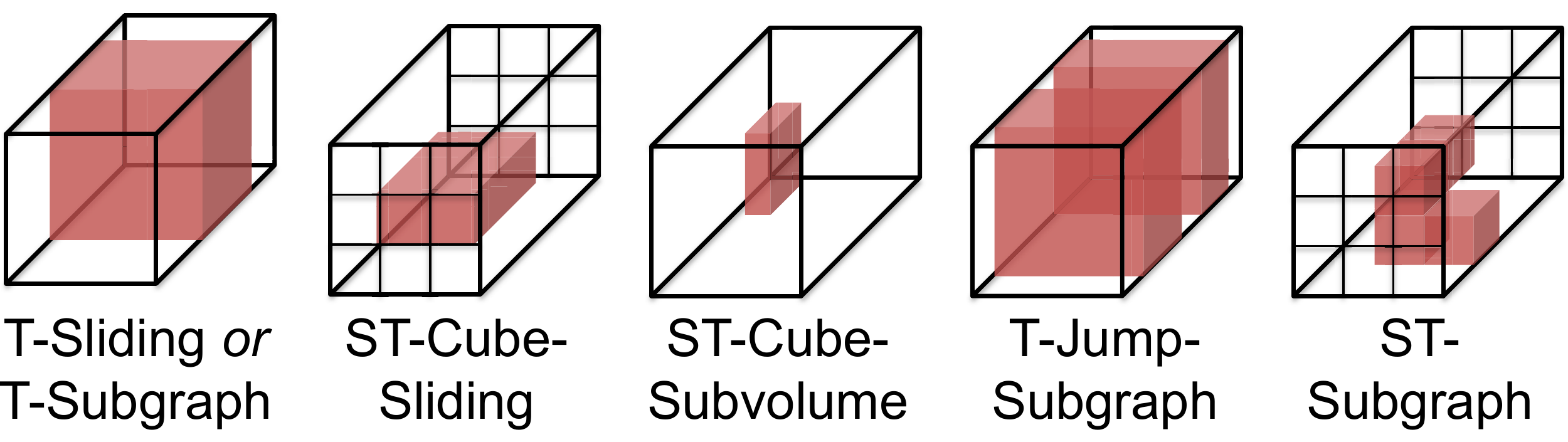}
   \caption{Sketch of the candidate subvolume types considered by different methods, ordered approximately from least to most flexible.  \textbf{T-Sliding or T-Subgraph}: The status quo sliding window search (and the proposed T-Subgraph without jumps) finds the full-frame subvolume believed to contain the activity (leftmost image).  \textbf{ST-Cube-Sliding}: A variant that performs sliding window on different spatial portions of the frame, with the restriction of cuboid subvolumes.   \textbf{ST-Cube-Subvolume}: A branch-and-bound search strategy from existing work~\cite{Yuan:2009:subvolumesearch} that considers all possible cube-shaped subvolumes---not just the grid-based subset considered by ST-Cube-Sliding.  \textbf{T-Jump-Subgraph}: The proposed method using temporal nodes (slabs of frames) only, with additional allowance of temporal ``gap(s)" in the output detections.  \textbf{ST-Subgraph} The most general form of the proposed method, where we use both spatial and temporal nodes, allowing irregular, non-cubic detection results.}
\label{fig:shapes}
\end{figure}

\section{Experimental Results}\label{sec:Experiment Result}

We next present experimental results applying our method for activity detection on several public benchmark datasets.  We evaluate our approach compared to both sliding window and sliding cuboid baselines as well as an existing state-of-the-art subvolume detection method that exploits branch-and-bound search.  Throughout we are interested in both the speed and accuracy attainable.  Ideally, we would like to achieve very accurate detection but at a small fraction of the run-time cost incurred by traditional sliding window methods.  Furthermore, in some scenarios we hope to improve the accuracy over sliding windows, since our method will permit searching a more complete set of windows than is tractable with a naive search implementation.

In what follows, we first describe the datasets, baselines, and metrics used in our experiments, and we provide implementation details for our approach not already covered above.  Then, the next four subsections present results organized around each of the four datasets.  This is the most natural organization, since the dataset properties and their respective available ground truth dictate which variants of our approach are relevant for testing (e.g., temporal detection only, fully spatio-temporal, two-stage for spatio-temporal with long sequences, etc.).

\subsubsection{Activity Detection Datasets}

We validate on four datasets, all of which are publicly available:
\begin{itemize}
\item \textbf{UCF Sports}~\cite{Rodriguez:2008:ucfdata}\footnote{\texttt{http://crcv.ucf.edu/data/UCF\_Sports\_Action.php}}:  UCF Sports consists of 10 actions from various sports typically found on TV, such as diving, golf swing, running, and skate boarding.  The data originates from stock footage websites like BBC Motion or GettyImages.  The provided clips are trimmed to the action of interest, so we expand them into longer test sequences by concatenating clips to form ``UCF-Concat" (details below).  The ground truth contains the action label and the bounding box annotation of the human.

\item \textbf{Hollywood Human Actions}~\cite{Laptev:2008:hollywood}\footnote{\texttt{http://www.di.ens.fr/~laptev/actions/}}: The training set contains 219 clips originating from 12 Hollywood movies, and the test set contains 211 clips from a disjoint set of 20 Hollywood movies.  The activities are things like answer phone, get out of car, shake hands, etc.  We test with the noisy ``uncropped" versions of the test sequences which are only roughly aligned with the action and contain about 40\% extraneous frames.  In all data there is a variety of camera motion and dynamic scenes.  The ground truth consists of the action label for the clip, as well as the correct temporal boundaries of the activity in the case of the uncropped sequences.

\item \textbf{MSR Actions}~\cite{Yuan:2009:subvolumesearch}\footnote{\texttt{http://research.microsoft.com/en-us/um/people/ \\ \;   zliu/actionrecorsrc/}}: The MSR dataset consists of 16 test clips with three activity classes---hand clapping, hand waving, and boxing---performed in front of cluttered and moving backgrounds.  They are performed by 10 subjects, both indoor and outdoor.  The ground truth consists of a spatio-temporal bounding box for each action.  To our knowledge, this is the only available activity dataset with both spatial and temporal annotations (others are limited to temporal boundaries only).  For this dataset, we train the activity classifiers using the disjoint KTH dataset~\cite{Laptev:2004:recognizinghumanactions}, following~\cite{Yuan:2009:subvolumesearch}.

\item \textbf{THUMOS 2014}~\cite{THUMOS14}\footnote{\texttt{http://crcv.ucf.edu/THUMOS14/}}: THUMOS consists of videos collected from YouTube containing 101 different action classes.  The emphasis on the THUMOS challenge is to cope with temporally untrimmed videos.  Accordingly, the test sequences contain the target actions naturally embedded in other content, and the ground truth includes the temporal boundaries of the true action.  Following the localization setting of the winners for the ECCV 2014 workshop's detection task~\cite{thumos:LEAR}, we divide the 1010 validation videos into two equal parts for testing and training.  The test data contains 20 activity classes: baseball pitch, basketball dunk, billiards, clean and jerk, cliff diving, cricket bowling, cricket shot, diving, frisbee catch, golf swing, hammer throw, high jump, javelin throw, long jump, pole vault, shot put, soccer penalty, tennis swing, throw discus, volleyball spiking.

\end{itemize}

See Table~\ref{table:dataset} for a summary of the dataset properties.  In particular, we include each dataset's typical clip lengths and the portion of the sequence occupied by the action to be detected.   On average, the action of interest occupies only 28\% of the total test sequence, making detection (as opposed to classification) necessary.

\begin{table}[t]
\scriptsize
\centering

\begin{tabular}{|p{1.3cm}|p{1.75cm}|p{0.7cm}|p{0.95cm}|p{0.95cm}|}
\hline
Dataset & Features & Num test videos & Ave length (\#frames) & Ave length of action \\
\hline
UCF-Concat & Dense$+$HoG3D & 12 & 589 & 13\% \\
\hline
Hollywood  & STIP+HoG/HoF & 211 & 474 & 62\%  \\
uncropped & or high-level & & & \\
\hline
MSR Action & STIP+HoG/HoF & 16 & 756 & 10\%  \\
\hline
THUMOS & STIP+HoG/HoF, Trajectory, MBH & 111 & 1717 & 29\%  \\
\hline
\end{tabular}

\caption{Properties of the four datasets.  See text for more details.}
\label{table:dataset}
\end{table}

\subsubsection{Baselines}

We compare our approach to three baselines:

\begin{itemize}
\item \textbf{T-Sliding}: a standard temporal sliding window.  This is the status quo method in the literature, e.g.,~\cite{Ke:2005:volumetricfeatures,Duchenne:2009:automaticannotation,Satkin:2010:modelingtemporalextent}.  Its results are equivalent to our T-Subgraph variant (using temporal linking structure), but computed with exhaustive search.

\item \textbf{ST-Cube-Sliding}: a variant of sliding window that searches all cuboid subvolumes having any \emph{rectangular} combination of the spatial-nodes used by our method.  Its search scope is similar to our ST-Subgraph, \emph{except} that it lacks all possible spatial links, meaning the detected subvolume cannot shift spatial location over time.  While most existing methods simply apply a sliding temporal window, with no spatial localization, we include this baseline as the natural straightforward extension of sliding window search if one wants to obtain localization.

\item \textbf{ST-Cube-Subvolume}: the state-of-the-art branch-and-bound method of~\cite{Yuan:2009:subvolumesearch}.  It considers \emph{all possible} cube-shaped subvolumes, and returns the one maximizing the sum of feature weights inside.  Its scope is more flexible than ST-Cube-Sliding.  Its objective is identical to ours, \emph{except} that it is restricted to searching cube-shaped volumes that cannot shift spatial location over time.  We use the authors' code.\footnote{We found its behavior sensitive to its \emph{penalty value} parameter, which is a negative prior on zero-valued pixels~\cite{Yuan:2009:subvolumesearch}.  The default setting was weak for our data, so for fairest comparisons, we tuned for best results on UCF.}
\end{itemize}

We stress that our approach is a new strategy for \emph{detection}; results in the literature focus largely on \emph{classification}, and so are not directly comparable.  The sliding window and subvolume baselines are state-of-the-art methods for detection, so our comparisons \emph{with identical features and classifiers} will give clear insight into our method's performance.

We consider four variants of our approach: T-Subgraph, T-Jump-Subgraph, ST-Subgraph, and two-stage ST-Subgraph, as defined in Sec.~\ref{sec:Algorithm Sketch}.  Recall that T-Subgraph \emph{provides equivalent accuracy to T-Sliding, but is faster}.\footnote{For the special case of temporal search, one can obtain equivalent solutions using 1-D branch-and-bound search to detect the max subvector along the temporal axis~\cite{Bentley:1984:PPA}. In practice we find this method's run-time to be similar or slightly faster than T-Subgraph.  Note, however, that it is \emph{not} applicable for any other search scope handled by our approach.}  The other two variants, T-Jump-Subgraph and ST-Subgraph, provide more flexibility for detection compared to any of the above methods.   In particular, the T-Jump-Subgraph variant \emph{allows temporal discontinuities} not permitted by any of the above methods, and the ST-Subgraph variant \emph{allows spatial changes} where the detected content can move spatially within the frame over time.  The two-stage ST-Subgraph (cf.~Sec.~\ref{sec:two_stage}) is like the latter, only computed in an approximate form so as to scale well to longer test sequences.

Figure~\ref{fig:shapes} depicts the  scope of the regions searched by each method, both ours and the baselines.

\subsubsection{Evaluation Metrics}

We adopt standard metrics for detection evaluation.  Following~\cite{Yao:2010:houghvoting,Klaser:2010:humanfocusedactionlocalization,Yuan:2009:subvolumesearch}, we use the \emph{mean overlap accuracy}.  Whether performing temporal or full spatio-temporal detection, this metric computes the intersection of the predicted detection region with the ground truth, divided by the union.  We use detection time (on our 3.47GHz Intel Xeon CPUs) to evaluate computational cost.

\subsubsection{Implementation Details}

For all datasets, we train a binary SVM to build a detector for each action.  We use the descriptors described in Sec.~\ref{sec:features}, following the guidance of prior work~\cite{WUKLS:2009:bmvcrecognition,wang:2011:dense-trajectory} to select which particular sampling strategies and local space-time descriptors to employ per dataset.  In particular, recommendations from~\cite{WUKLS:2009:bmvcrecognition} lead us to employ HoG/HoF for Hollywood and HoG3D for UCF with dense sampling.  For the THUMOS dataset we use the features provided with the dataset, which augments the HoG/HoF set with dense trajectories and MBH.  In particular, on THUMOS we train one-versus-all binary SVMs with four types of features: trajectory~\cite{wang:2011:dense-trajectory}, HOG, HOF, and MBH~\cite{oneata:2013:mbh}, where the features are quantized to a bag of words representation via k-means with a dictionary size $=4000$.  We use the authors' code for HoG3D/HoG/HoF/trajectory/MBH~\cite{Laptev:2008:hollywood,kalser:2008:HoG3D,wang:2011:dense-trajectory,oneata:2013:mbh}, with default parameter settings.  We test the high-level descriptors on Hollywood, since that dataset has substantial person-object interactions, whereas actions in the others are more person-centric (e.g., diving, clapping, skateboarding).   We construct our temporal graphs with a node size of 10 frames per slab.

The next four sections describe the results on each dataset in turn.

\subsection{Temporal Detection on UCF Sports}\label{sec:ucf_result}

\begin{table}[t]
\scriptsize
\centering
\resizebox{0.5\textwidth}{!}{
\begin{tabular}{|l|p{1.1cm}|p{1.1cm}|p{1.1cm}|p{1.3cm}|}
\hline
Verbs & T-Sliding & ST-Cube-Subvol~\cite{Yuan:2009:subvolumesearch} & Our-T-Subgraph & Our-T-Jump-Subgraph \\
\hline
Diving & 0.8106 & 0.7561 & 0.8106 & \textbf{0.9091} \\
\hline
Lifting & 0.7899 & 0.8058 & 0.7899 & \textbf{0.8096} \\
\hline
Riding & \textbf{0.5349} & 0.5075 & \textbf{0.5349} & 0.3888 \\
\hline
Running & 0.4602 & 0.3269 & 0.4602 & \textbf{0.4705} \\
\hline
Skateboard & 0.1407 & 0.1057 & 0.1407 & \textbf{0.1803} \\
\hline
Swing-Bench & 0.5520 & \textbf{0.6259} & 0.5520 & 0.4582 \\
\hline
Swing-Side & 0.6728 & 0.3478 & 0.6728 & \textbf{0.7212} \\
\hline
Walking & 0.4085 & 0.3462 & 0.4085 & \textbf{0.4657} \\
\hline
\end{tabular}
}
\caption{Mean overlap accuracy for the UCF Sports data.}
\label{table:ucf_overlapping}
\end{table}

\begin{table}[t]
\scriptsize
\centering
\begin{tabular}{|p{1.4cm}|p{1.2cm}|p{1.2cm}|p{1.2cm}|p{1.3cm}|}
\hline
Detection time (ms) & T-Sliding & ST-Cube-Subvol~\cite{Yuan:2009:subvolumesearch} & Our-T-Subgraph & Our-T-Jump-Subgraph \\
\hline
Mean & $1.25\times10^5$ & $7.87\times10^4$ & \textbf{$1.02\times10^2$} & $6.51\times10^2$ \\
\hline
Stdev & $7.52\times10^3$ & $3.17\times10^4$ & \textbf{$5.35\times10^1$} & $3.17\times10^2$ \\
\hline
\end{tabular}
\caption{Search time for the UCF Sports data.}
\label{table:ucf_time}
\end{table}

Since the UCF clips are already cropped to the action of interest, we modify it to make it suitable for detection.  We form 12 test sequences by concatenating 8 different clips each from different verbs.  All test videos are totally distinct, and are available on our project website.  We train the SVM on a disjoint set of cropped instances.  We perform temporal detection only, since the activities occupy the entire frame.

\begin{figure*}[t]
\centering
\includegraphics[width=0.7\textwidth]{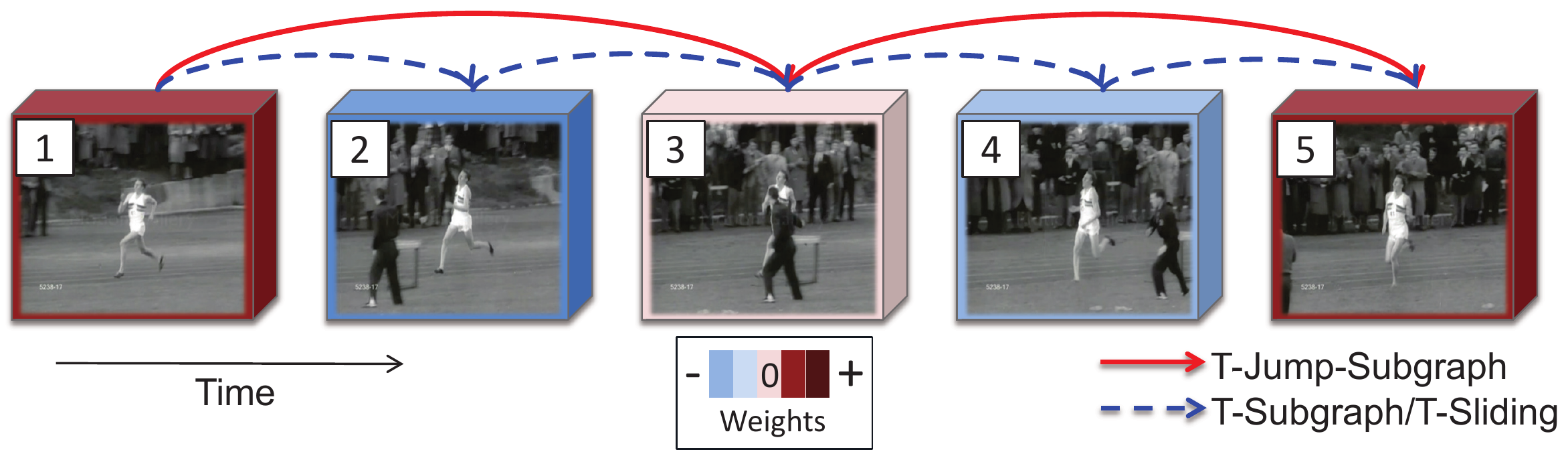}
   \caption{Qualitative example showing how our T-Jump method can perform robust detection.  The five colored cubes represent the weighted node computed from the extracted features and learned classifier. For the second to fourth nodes, the classifier generates negative weights due to the occlusion. Using T-Sliding or T-Subgraph, the detection output does not cover the first and last cubes due to the negative weights from three cubes in the middle.  In contrast, using our T-Jump method, it can skip over the intervening negative weights.  This makes the detection framework more robust to noise from occlusion. Best viewed in color.}
\label{fig:ucf_qualitative}
\end{figure*}

Table~\ref{table:ucf_overlapping} shows the accuracy results, and Table~\ref{table:ucf_time} shows the search times.  For almost all verbs, our subgraph approaches outperform the baselines.  Further, our T-Jump variant gives top accuracy in most cases, showing the advantage of ignoring noisy features (in this data, often found near the onset or ending of the verb).   Figure~\ref{fig:ucf_qualitative} shows an example where T-Jump performs robust detection in spite of occlusions, whereas the baseline sliding window or basic T-Sliding fails.

On this dataset, the ST-Cube-Subvolume baseline is often weaker than sliding window.  Upon inspection, we found it often fires on a small volume with highly weighted features when the activity changes in spatial location over time.  However, it is best on ``Swing-Bench'', likely because the backgrounds are fairly static, minimizing misleading features.  As we see in Table~\ref{table:ucf_time}, both our subgraph methods are orders of magnitude faster than the baselines.  Note that the ST-Cube-Subvolume's higher cost is reasonable since here it is searching a wider space.

\subsection{Temporal Detection on Hollywood}\label{sec:hollywood_result}

\begin{table}[t]
\scriptsize
\centering
\resizebox{0.5\textwidth}{!}{
\begin{tabular}{|l|p{1.1cm}|p{1.1cm}|p{1.1cm}|p{1.3cm}|}
\hline
Verbs & T-Sliding & ST-Cube-Subvol~\cite{Yuan:2009:subvolumesearch} & Our-T-Subgraph & Our-T-Jump-Subgraph \\
\hline
AnswerPhone & 0.3968 & 0.2905 & 0.3968 & \textbf{0.3994} \\
\hline
GetOutCar & 0.2276 & 0.2267 & 0.2276 & \textbf{0.2921} \\
\hline
HandShake & 0.3071 & 0.3390 & 0.3071 & \textbf{0.3663} \\
\hline
HugPerson & 0.3869 & \textbf{0.4486} & 0.3869 & 0.4150 \\
\hline
Kiss & 0.3822 & 0.4230 & 0.3831 & \textbf{0.4412} \\
\hline
SitDown & \textbf{0.3612} & 0.2861 & \textbf{0.3612} & 0.3550 \\
\hline
SitUp & 0.2592 & 0.2053 & 0.2592 & \textbf{0.3255} \\
\hline
StandUp & 0.3475 & 0.3013 & 0.3475 & \textbf{0.3775} \\
\hline
\end{tabular}
}
\caption{Mean overlap accuracy on uncropped Hollywood data.}
\label{table:hollywood_overlapping}
\end{table}

\begin{table}[t]
\scriptsize
\centering
\begin{tabular}{|p{1.4cm}|p{1.2cm}|p{1.2cm}|p{1.2cm}|p{1.3cm}|}
\hline
Detection Time (ms) & T-Sliding & ST-Cube-Subvol~\cite{Yuan:2009:subvolumesearch} & Our-T-Subgraph & Our-T-Jump-Subgraph \\
\hline
Mean & $3.71\times10^3$ & $1.70\times10^5$ & \textbf{$6.63\times10$} & $5.69\times10^2$ \\
\hline
Stdev & $1.03\times10^4$ & $5.79\times10^5$& \textbf{$7.51\times10$} & $1.77\times10^3$ \\
\hline
\end{tabular}
\caption{Search time on uncropped Hollywood data.}
\label{table:hollywood_time}
\end{table}

We next test the Hollywood data, which also permits a study of temporal detection.  As noted above, we test with the untrimmed data provided by the dataset creators.  Existing work uses this data for classification, and so trains \emph{and} tests with the cropped versions.  To perform temporal detection, we instead train with the cropped clips, and test with the uncropped clips.

Table~\ref{table:hollywood_overlapping} shows the accuracy results, and Table~\ref{table:hollywood_time} shows the search times.  Our T-Jump-Subgraph achieves the best accuracy for 6 of the 8 verbs, with even more pronounced gains than on UCF.  This again shows the value of skipping brief negatively weighted portions; e.g., ``AnswerPhone'' can transpire across several shot boundaries, which tends to mislead the baselines.

As Table~\ref{table:hollywood_time} reveals, our method is again significantly faster than the baselines.  Our T-Jump-Subgraph is slower than our T-Subgraph search, given the higher graph complexity (which also makes it more accurate).  Hence, which variant to apply depends on how an application would like to make this cost-accuracy trade off.

\begin{table}[t]
\footnotesize
\centering
\begin{tabular}{|l|c|}
\hline
Test sequence composition & Accuracy \\
\hline
Raw uncropped clips & 24.83\% \\
\hline
Output from T-Subgraph & 29.66\% \\
\hline
Manual ground truth & 29.97\% \\
\hline
\end{tabular}
\caption{Recognition accuracy on Hollywood as test input varies.}
\label{table:hollywood_accuracy}
\end{table}

One might wonder whether a naive detector that simply classifies the entire uncropped clip could do as well.  To check, we compare \emph{recognition} results when we vary the composition of the test sequence to be either (a) the uncropped clip, (b) the output of our detector, or (c) the ground truth cropped clip.  Table~\ref{table:hollywood_accuracy} shows the result.  We see indeed that detection is necessary; using our output is much better than the raw untrimmed clips, and only slightly lower than using the manually provided ground truth.

We also test our high-level descriptor (cf.~Sec.~\ref{sec:high}) on Hollywood, since its actions contain human-object interactions.  We apply six object detectors---bus, car, chair, dining table, sofa, and phone---to every fifth frame, and use random forests with 10 trees. Table~\ref{table:hollywood_high_level_feat} shows the results, compared to our method using low-level features.  For five of the eight actions, the proposed high-level descriptor improves accuracy.  It is best for activities based on the interaction between two people (e.g., kiss) or involving an obvious change in pose (e.g., sit up), showing the strength of the proposed person types to capture pose and temporal ordering.  For other verbs with varied objects (answer phone, get out of car), it hurts accuracy, likely due to object detector failures in this dataset.  It remains future work outside the scope of this project to bolster the component object detectors fed into this higher-level neighborhood descriptor.

\begin{table}[t]
\scriptsize
\centering
\begin{tabular}{|l|c|c|}
\hline
Verbs & T-Subgraph (HoG/HoF) & T-Subgraph (high-level) \\
\hline
AnswerPhone & \textbf{0.3968} & 0.1741 \\
\hline
GetOutCar & \textbf{0.2276} & 0.1447 \\
\hline
HandShake & 0.3071 & \textbf{0.4194} \\
\hline
HugPerson & 0.3869 & \textbf{0.5292} \\
\hline
Kiss & 0.3822 & \textbf{0.4906} \\
\hline
SitDown & 0.3612 & \textbf{0.3753} \\
\hline
SitUp & 0.2592 & \textbf{0.3843} \\
\hline
StandUp & \textbf{0.3475} & 0.2636 \\
\hline
\end{tabular}
\caption{Mean overlap accuracy on Hollywood for low-level features vs.~the object-based high-level descriptors.}
\label{table:hollywood_high_level_feat}
\end{table}

\subsection{Temporal Detection with Multiple Instances on THUMOS}

Next we evaluate our approach on the THUMOS dataset.  THUMOS allows temporal detection (like UCF Sports and Hollywood), plus, unlike the others, it contains test sequences with multiple instances of the activity.  This aspect lets us test our iterative max-subgraph strategy to produce multiple detections, as discussed in Sec.~\ref{sec:multiple_instance_detection}.

In these experiments, the sliding window baseline represents the same search strategy taken by the leading approach on this dataset~\cite{thumos:LEAR}.  As such, we follow the authors' parameter choices for the window search in order to provide a close comparison.  That means for the T-Sliding baseline, we use a step size of 10 frames, and evaluate the windows with durations of 10, 20, 30, 40, 50, 60, 70, 80, 90, 100, and 150 frames~\cite{thumos:LEAR}.  We fix the NMS threshold at 0.5 (after we did not observe better results for the baseline shifting this threshold within the range (0,1]), and we fix the node re-weighting value at 0 for our method (cf.~Sec.~\ref{sec:multiple_instance_detection}).   Note that with a skip size of 10 frames, the sliding window baseline (T-Sliding) does not exhaustively search all subsequences, whereas our method does.  For each testing video, we return up to 10 positive detection windows.

Table~\ref{table:thumos_accuracy} shows the accuracy results for T-Sliding and our T-Subgraph method, both in terms of overlap and the mean average precision (mAP) as defined by~\cite{THUMOS14}, which is a useful metric for the case when there are multiple instances per testing clip.  Our method obtains higher accuracy than the standard sliding window baseline.  This is a direct consequence of the efficiency of our approach in considering all possible windows.  We also get a noticeable further advantage in overlap accuracy applying our T-Jump variant, yet it harms average precision.  Upon inspection, we find that for this challenging data, the  classifier scores per node are noisier, which leads T-Jump to cover too many frames; T-Jump can easily find some small-valued positive nodes to skip over highly negative nodes, leading to some poorer detection outputs as seen in the mAP. The high overlapping score of T-Jump confirms this observation and illustrates why mAP is a better metric than overlapping accuracy in multiple instance detection. We also tried a variant of our approach that less aggressively reduces the weights on nodes already involved in a prior iteration's detections: we set the weight of a ``used" node to the mean weight of all nodes, with the intent to encourage more overlapping detections.  However, this led to slightly worse accuracy for our method (0.2043 overlap accuracy vs.~0.2186 in Table~\ref{table:thumos_accuracy}).

Table~\ref{table:thumos_time} shows the computation time for both methods.  Similar to previous results, our T-Subgraph method for detecting multiple instances provides significantly faster running time compared to T-Sliding.  For the sliding window method, no matter how many output detections we want, all the candidate window are evaluated. In contrast, for our T-Subgraph, we only return one optimal window in each subgraph search iteration and re-weight the underlying nodes for next iteration.  Therefore, in this experiment, we need to run our T-Subgraph 10 times to find top 10 detection windows---yet, in spite of that repetition, it is still about an order of magnitude faster than evaluating all the candidate windows in the T-Sliding method.

\begin{table}[t]
\footnotesize
\centering
\begin{tabular}{|l|c|c|c|}
\hline
Metric & T-Sliding & Our T-Subgraph & Our T-Jump-Subgraph\\
\hline
mAP      & 0.1983 & \textbf{0.2143} & 0.1546\\
\hline
Overlap & 0.1792 & 0.2186 & \textbf{0.2636}\\
\hline
\end{tabular}
\caption{Recognition accuracy on THUMOS 2014 data.}
\label{table:thumos_accuracy}
\end{table}

\begin{table}[t]
\scriptsize
\centering
\begin{tabular}{|l|c|c|c|}
\hline
Time (ms) & T-Sliding & Our T-Subgraph & Our T-Jump-Subgraph \\
\hline
Mean & $7.07\times10^5$ & $5.34\times10^4$ & $4.72\times10^4$\\
\hline
Stdev & $2.26\times10^6$ & $2.37\times10^5$ & $1.97\times10^5$\\
\hline
\end{tabular}
\caption{Search time on THUMOS 2014 data.}
\label{table:thumos_time}
\end{table}

Finally, we more closely analyze the behavior of the sliding window baseline (T-Sliding) as it compares to our T-Subgraph.  The goal is to see in practice what density of windowed search (skip sizes) is necessary for best results.  In other words, if we allow T-Sliding more candidate windows and hence longer running time, at what point does it come close to the optimal result from our method?  Since running this experiment is rather costly for the baseline, we limit this test to four of the 20 verbs in the THUMOS test set (chosen randomly: basketball dunk, clean and jerk, cliff diving, and hammer throw).

\begin{figure}[t]
\centering
\includegraphics[width=0.35\textwidth]{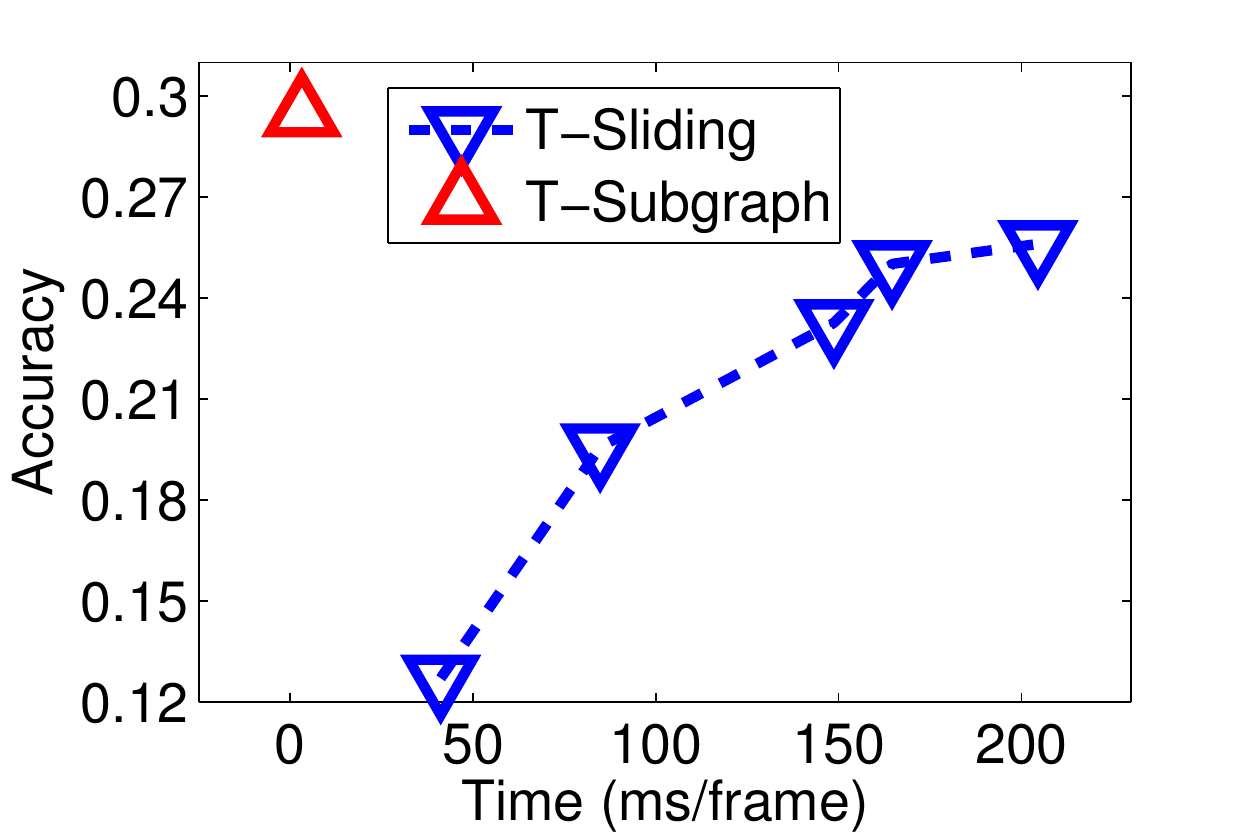}
   \caption{Accuracy vs.~computation time in temporal search.  We compare our T-Subgraph (which produces the optimal detection output for a fixed time) to the standard T-Sliding method (which produces its detection output based on exhaustive search of a pool of candidate windows).  Here we increase T-Sliding's accuracy and run-time by increasing that pool of windows.}
\label{fig:time_accuracy_curve}
\end{figure}

Figure~\ref{fig:time_accuracy_curve} shows the results in terms of the average accuracy over all four actions tested.  As expected, increasing the pool of candidate windows searched by T-Sliding increases its accuracy, but at a corresponding linear increase in run-time.  At a search time of 200 \emph{ms} per frame, the baseline is searching 35 different window sizes (out of ~300 window sizes for exhausted search) and achieves accuracy of 0.26, nearing but not as good as the result from T-Subgraph of 0.30 accuracy obtained with just a few \emph{ms} per frame.

\subsection{Space-Time Detection on MSR Actions}\label{sec:msr_result}

As the fourth and final dataset, we experiment with MSR Actions.  In contrast to all of the above datasets, MSR Actions contains ground truth for the \emph{spatial} localization of the action---not just the temporal extent.  Furthermore, the actors change their position over time and a test sequence may contain multiple simultaneous instances of different actions.  Therefore, this dataset is a good testbed to evaluate our ST-Subgraph with the node structure in Figure~\ref{fig:node_formulation}(b), where we link neighboring nodes both in space and time.   In what follows, we present results with both the exact maximum subgraph from ST-Subgraph as well as its approximate counterpart, the two-stage search process described in Sec.~\ref{sec:two_stage}.

\begin{figure*}[t]
\centering
\includegraphics[width=0.7\textwidth]{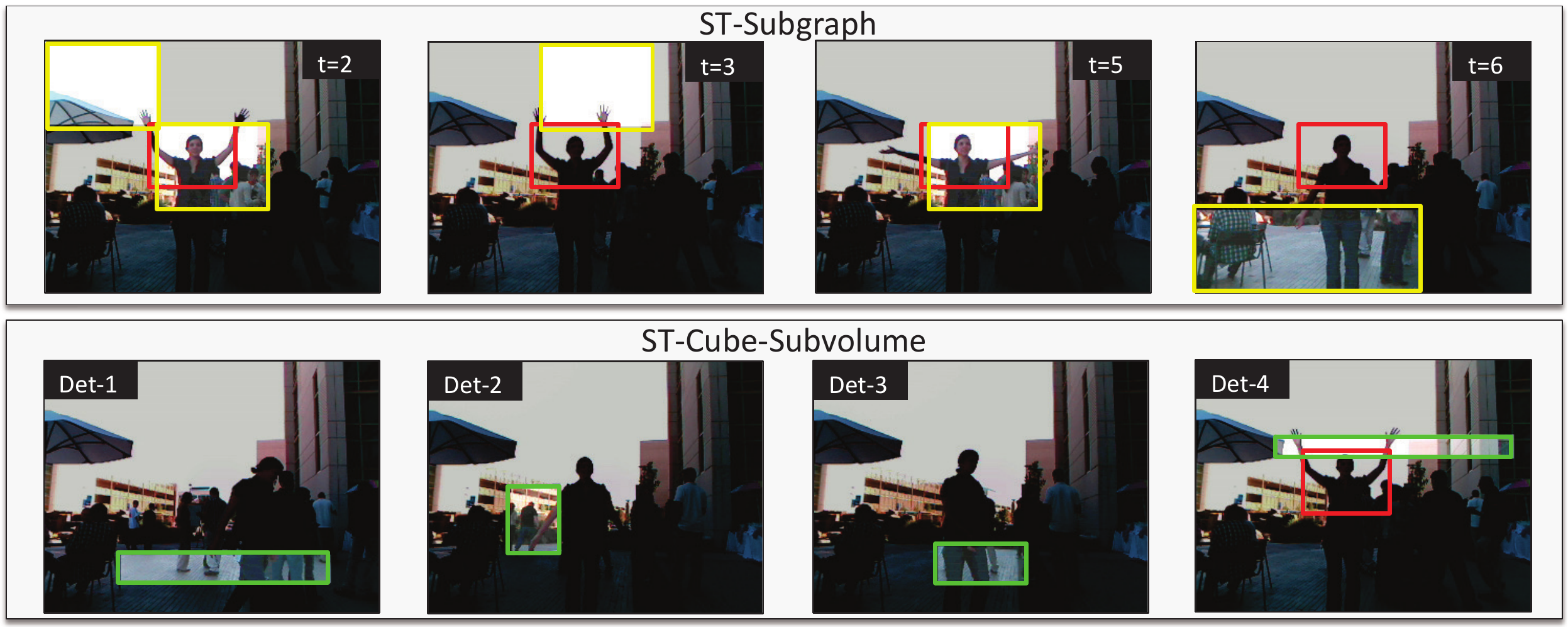}
   \caption{Example of ST-Subgraph's top output (top) and the top 4 detections from ST-Cube-Subvolume~\cite{Yuan:2009:subvolumesearch} (bottom).  Red rectangles denote ground truth.  Brighter areas denote detections.}
\label{fig:qualitative_result_msr}
\end{figure*}

First we isolate temporal detection accuracy alone.  We run the temporal and spatio-temporal variants of our method, and project the spatio-temporal results to temporal results.  Table~\ref{table:msr_overlapping} shows results.  Even under the temporal criterion, our ST-Subgraph and two stage ST-Subgraph are most accurate, since they can isolate those nodes that participate in the action.  Figure~\ref{fig:qualitative_result_msr} illustrates how our space-time node structure succeeds when the location of activity changes over time, whereas ST-Cube-Subvolume may be trapped in cube-shaped maxima. Compared to ST-Subgraph, our two-stage method yields similar accuracy for Boxing and Clapping videos and provides lower accuracy for Waving videos. This result shows the two-stage method is able to provide good approximation to ST-subgraph method.

\begin{table}[t]
\scriptsize
\centering
\resizebox{0.5\textwidth}{!}{
\hspace*{-0.10in}
\begin{tabular}{|p{1.05cm}|p{1.05cm}|p{1.05cm}|p{1.05cm}|p{1.05cm}|p{1.05cm}|p{1.05cm}|}
\hline
Verbs & T-Sliding & ST-Cube-Sliding & ST-Cube-Subvol~\cite{Yuan:2009:subvolumesearch} & Our-T-Subgraph & Our-ST-Subgraph & Our-Two-Stage-ST\\
\hline
Boxing & 0.0541 & 0.0717 & 0.0794 & 0.0541 & 0.0989 & \textbf{0.1188}\\
\hline
Clapping & 0.0982 & 0.0982 & 0.0602 & 0.0982 & 0.1754 & \textbf{0.1795}\\
\hline
Waving & 0.2342 & 0.2204 & 0.2669 & 0.2342 & \textbf{0.2926} & 0.2416\\
\hline
\end{tabular}
}
\caption{Mean temporal overlap accuracy on the MSR dataset.}
\label{table:msr_overlapping}
\end{table}

\begin{table}[t]
\scriptsize
\centering
\resizebox{0.5\textwidth}{!}{
\begin{tabular}{|p{1.2cm}|p{1.05cm}|p{1.05cm}|p{1.05cm}|p{1.05cm}|p{1.05cm}|p{1.05cm}|}
\hline
Detection Time (ms) & T-Sliding & ST-Cube-Sliding & ST-Cube-Subvol~\cite{Yuan:2009:subvolumesearch} & Our-T-Subgraph & Our-ST-Subgraph & Our-Two-Stage-ST\\
\hline
Mean & $4.2\times10^3$ & $5.5\times10^4$ & $3.0\times10^5$ & \textbf{$2.8\times10^2$} & $3.1\times10^6$ & $1.4\times10^3$\\
\hline
Stdev & $3.3\times10^3$ & $4.2\times10^4$ & $1.6\times10^5$ & \textbf{$2.3\times10^2$} & $4.6\times10^6$ & $4.1\times10^2$\\
\hline
\end{tabular}
}
\caption{Search time on the MSR dataset.}
\label{table:msr_time}
\end{table}

\begin{table}[t]
\scriptsize
\centering
\resizebox{0.5\textwidth}{!}{
\hspace*{-0.10in}
\begin{tabular}{|p{1.7cm}|p{1.4cm}|p{1.4cm}|p{1.4cm}|p{1.4cm}|}
\hline
Verbs & ST-Cube-Sliding & ST-Cube-Subvol~\cite{Yuan:2009:subvolumesearch} & Our-ST-Subgraph & Our-Two-Stage-ST\\
\hline
Boxing & \textbf{0.0478} & 0.0193 & 0.0417 & 0.0296\\
\hline
Hand Clapping & 0.0373 & 0.0071 & \textbf{0.0630} & 0.0425\\
\hline
Hand Waving &  0.0851 & 0.0581 & \textbf{0.1121} & 0.0809\\
\hline
\end{tabular}
}
\caption{Mean space-time overlap accuracy on the MSR dataset.  (T-Sliding/T-Subgraph are omitted since they do not do spatial localization.)}
\label{table:msr_st_overlapping}
\end{table}

Next we examine the complete space-time localization accuracy.  Table~\ref{table:msr_st_overlapping} shows the results, evaluated under the ground truth annotation for the person who performs the action\footnote{The original ground truth labels only the hand regions (see Figure~\ref{fig:qualitative_result_msr}), whereas this ground truth labels the whole person performing the action.}.  Results are mixed between the methods, with a slight edge for our ST-Subgraph. Also, only the non-rectangular shape detection from our ST-Subgraph reflects the large spatial motions in actions.  As expected, the two-stage search process does detract from the accuracy of the optimal ST-Subgraph result, as we see in the last two columns of Table~\ref{table:msr_st_overlapping}.

Finally, we analyze the run-times for all methods tested in Table~\ref{table:msr_time}.  Here we see the substantial practical impact of our two-stage spatio-temporal variant, which yields significantly lower computation time.  It is even faster than the sliding temporal window search that produces no spatial localization, and orders of magnitude faster than the existing branch-and-bound subvolume method~\cite{Yuan:2009:subvolumesearch}.  The two-stage method is slightly slower than the T-Subgraph variant of our method, since it requires additional computation for the spatial detection in the first stage for each slab.

\begin{figure}[t]
\centering
\hspace*{-0.2in}
\includegraphics[width=8cm]{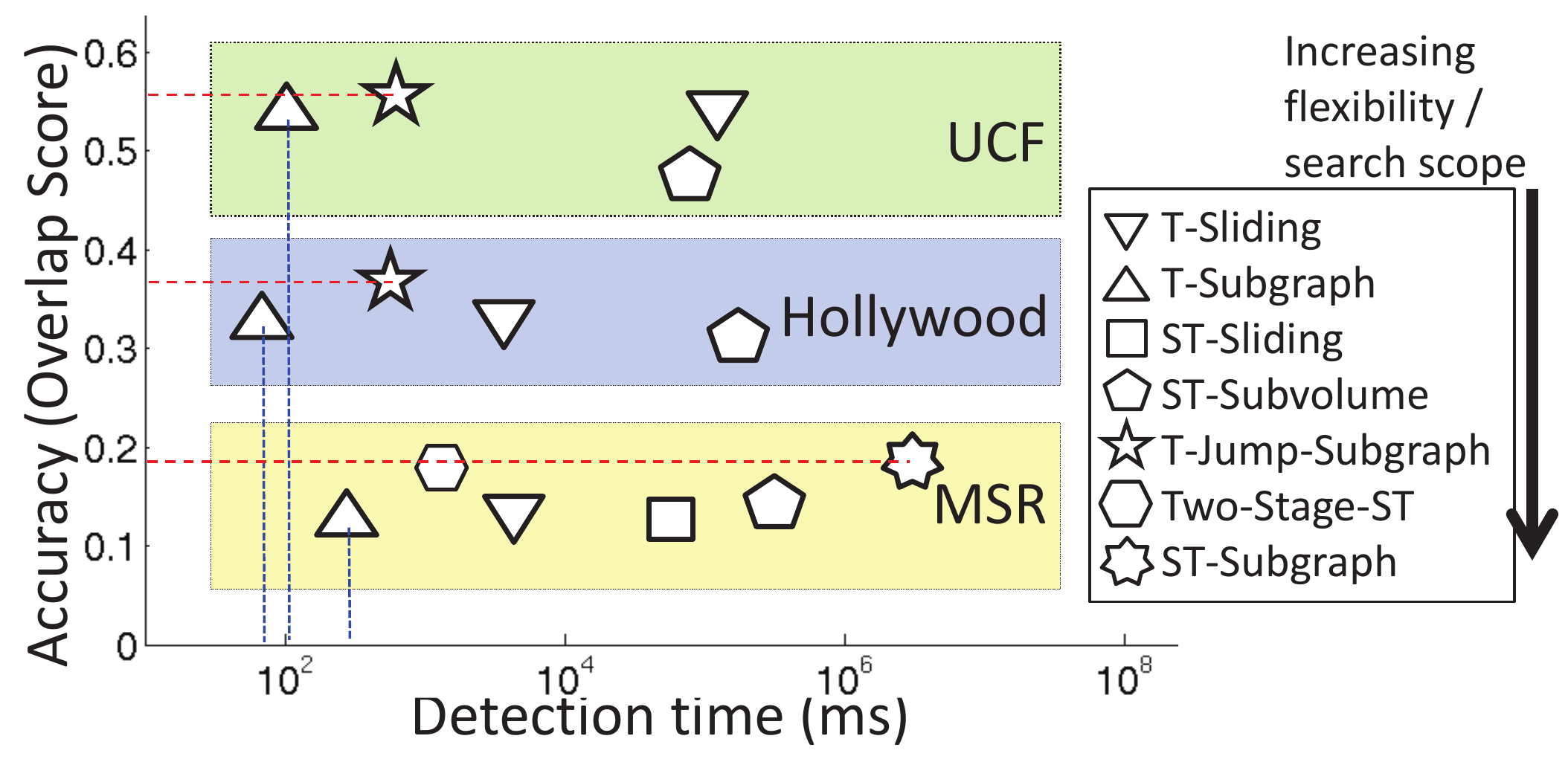}
   \caption{Overview of methods on the three datasets. } 
\label{fig:all_overlaping_time}
\end{figure}

As discussed in Sec.~\ref{sec:two_stage}, we can achieve efficient spatio-temporal localization with the our proposed two stage subgraph search method. In previous section, our ST-Subgraph provides more accurate space-time localization of action with higher computational cost. In this section, we speed up the ST-Subgraph with our two stage subgraph for space time detection on MSR action dataset.

Table~\ref{table:msr_st_overlapping} and Table~\ref{table:msr_time} also show the comparison of detection accuracy and search time for our Two-Stage-ST-Subgraph and our original ST-Subgraph. By dividing the node structure into temporal slices, the computation time of two stage method is reduced by three orders compared to original ST-Subgraph. As expected, the two stage method is slightly slower than the T-Subgraph because it requires additional computation for spatial detection in first stage for each temporal node. For detection accuracy, recall that the two stage method doesn't guarantee to provide the optimal spatial-temporal volumes since it ignores the temporal link between nodes in the first stage, it is expected that the two stage method will be less accurate than the ST-Subgraph method. As shown in Table~\ref{table:msr_st_overlapping}, Two-Stage-ST method achieves similar accuracy to the ST-Subgraph for hand clapping and hand waving clips, but lower accuracy for boxing clips. It is because the learned activity model for boxing is less accurate than the learned models for other two actions (it provides lower overlap accuracy for ST-Subgraph), and our two stage method is more sensitive to the noisy node score due to the pruned connections between nodes.

\subsection{Summary of Trade-Offs in Results}\label{sec:tradeoff}

Having presented all the results, now we step back and attempt to summarize the outcomes succinctly. There are three dimensions of trade-offs between all methods tested: search time, search scope, and detection accuracy.

Figure~\ref{fig:all_overlaping_time} summarizes all trade-offs for three datasets.  Here we show the accuracy versus the detection time for each result, and encode the search scope of the method by the complexity of its polygonal symbol.  More complex symbols mean wider search scope. For example, recalling Figure~\ref{fig:shapes}, the least complex search scope is T-Sliding/T-Subgraph, which is plotted as a triangle, whereas the most complex search scope is the ST-Subgraph, which is plotted as a 14-sided star.

Importantly, we see that increased search scope generally boosts accuracy.  In addition, the flexibility of the graph structure in our subgraph algorithm allows it to perform best per dataset in terms of \emph{either} speed (see vertical blue dotted lines) or accuracy (see horizontal red dotted lines).

Our method can be used to produce equivalent results as sliding window search, but without the exhaustive search.  However, due to the additive restriction our method places on the classifier (cf.~Sec.~\ref{sec:objective}), it cannot \emph{normalize} each window's bag-of-feature histograms.  Would such normalization help the accuracy of sliding windows?   We find it actually hurt the baseline, letting tiny subvolumes with few positively weighted features dominate the detection outputs.  We can improve the normalization by also re-weighting the detection score by the length of the window to encourage longer detections~\cite{thumos:LEAR}.  Table~\ref{table:normalization} shows the result.  The T-Sliding accuracy increases in three of the four datasets, yet remains inferior to our method's best results.  Our accuracy advantage comes from our flexible subgraph node and linking strategies.

\begin{table}[t]
\scriptsize
\centering
\resizebox{0.5\textwidth}{!}{
\begin{tabular}{|l|c|c||c|}
\hline
Dataset & T-Sliding & T-Sliding-Norm & Ours\\
\hline
UCF (ave. overlap)& \textbf{0.5453} & 0.5417 & \textbf{0.5504} \\
\hline
Hollywood (ave. overlap) & 0.3337 & \textbf{0.3565} & \textbf{0.3715} \\
\hline
MSR (ave. overlap) & 0.1288 & \textbf{0.1513} &  \textbf{0.1890} \\
\hline
THUMOS (ave. mAP) & 0.1983 & \textbf{0.2026} & \textbf{0.2143}  \\
\hline
\end{tabular}
}
\caption{Effect of histogram normalization and re-weighting for the sliding window baseline, compared to the best performing variant of our method (T-Jump-Subgraph on UCF, Hollywood, and MSR; T-Subgraph on THUMOS).}
\label{table:normalization}
\end{table}

We provide our source code and data in our project page.\footnote{\texttt{http://vision.cs.utexas.edu/projects/maxsubgraph}}

\section{Conclusions}
We presented a novel branch-and-cut subgraph framework for activity detection that efficiently searches a wide space of temporal or space-time subvolumes.  Compared to traditional sliding window search, it significantly reduces computation time.  Compared to existing branch-and-bound methods, its flexible node structure offers more robust detection in noisy backgrounds.  Our novel high-level descriptor also shows promise for complex activities, and makes it possible to preserve the spatio-temporal relationships between humans and objects in the video, while still exploiting the fast subgraph search.

 \section*{Acknowledgment}
We thank the anonymous reviewers for their feedback, and Sudheendra Vijayanarasimhan for helpful discussions.  This research is supported in part by ONR PECASE N00014-15-1-2291.

\ifCLASSOPTIONcaptionsoff
  \newpage
\fi

\bibliographystyle{ieee}
\bibliography{strings,ref}

\end{document}